\documentclass[10pt,twocolumn,letterpaper]{article}

\usepackage[pagenumbers]{iccv} 

\definecolor{iccvblue}{rgb}{0.21,0.49,0.74}
\usepackage[pagebackref,breaklinks,colorlinks,allcolors=iccvblue]{hyperref}
\usepackage{booktabs}
\usepackage{multirow}
\usepackage{cuted}
\newcommand{\datasetname}{AgroBench\xspace}
\def\paperID{3404} 
\def\confName{ICCV}
\def\confYear{2025}

\title{\datasetname: Vision-Language Model Benchmark in Agriculture}

\author{\large Risa Shinoda\textsuperscript{\rm 1,2,4},
Nakamasa Inoue\textsuperscript{\rm 3,4},
Hirokatsu Kataoka\textsuperscript{\rm 4,5},
Masaki Onishi\textsuperscript{\rm 4},
Yoshitaka Ushiku\textsuperscript{\rm 6}
\\
\\
\textsuperscript{\rm 1}The University of Osaka
\textsuperscript{\rm 2}Kyoto University
\textsuperscript{\rm 3}Tokyo Institute of Technology\\
\textsuperscript{\rm 4}National Institute of Advanced Industrial Science and Technology (AIST)\\
\textsuperscript{\rm 5}Visual Geometry Group, University of Oxford
\textsuperscript{\rm 6}OMRON SINIC X
}

\begin{document}
\maketitle
\begin{strip}\centering
\vspace{-0.5in}
\includegraphics[width=\textwidth]{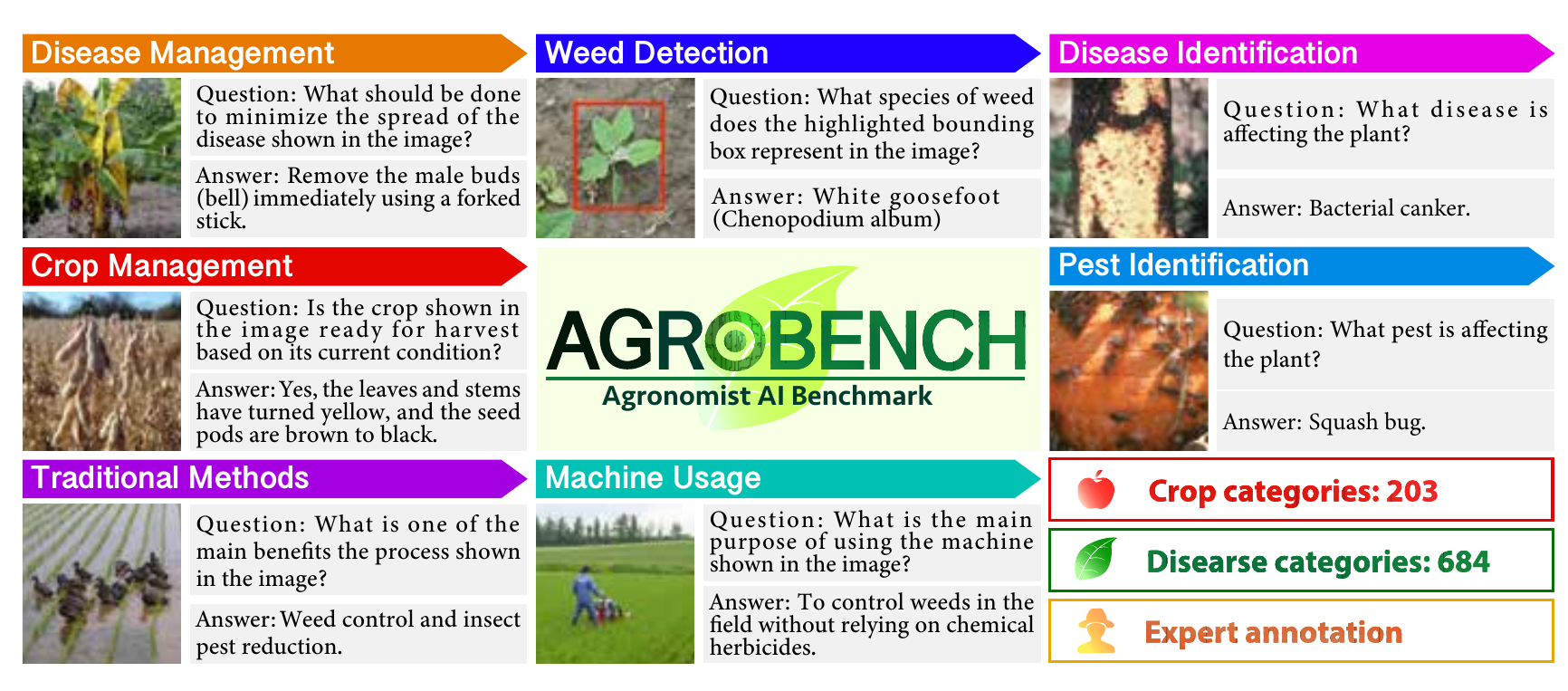}
 \vspace{-0.3in}
\captionof{figure}{
We present \datasetname~(\textbf{Agro}nomist AI \textbf{Bench}mark) designed to comprehensively evaluate 682 disease categories across 203 agricultural crop types for 7 vision-language question-answer tasks. In the era of larger-scale vision-language models (VLMs), our \datasetname is obviously non-trivial in terms of many more crop and disease categories with all expert annotations for establishing QA benchmarks in the agricultural domain.}

 \label{fig:fig1}
\end{strip}
\begin{abstract}
Precise automated understanding of agricultural tasks such as disease identification is essential for sustainable crop production.
Recent advances in vision-language models (VLMs) are expected to further expand the range of agricultural tasks by facilitating human-model interaction through easy, text-based communication.
Here, we introduce \datasetname~(\textbf{Agro}nomist AI \textbf{Bench}mark), a benchmark for evaluating VLM models across seven agricultural topics, covering key areas in agricultural engineering and relevant to real-world farming.
Unlike recent agricultural VLM benchmarks, \datasetname~is annotated by expert agronomists.
Our \datasetname~covers a state-of-the-art range of categories, including 203 crop categories and 682 disease categories, to thoroughly evaluate VLM capabilities.
In our evaluation on \datasetname, we reveal that VLMs have room for improvement in fine-grained identification tasks. Notably, in weed identification, most open-source VLMs perform close to random. With our wide range of topics and expert-annotated categories, we analyze the types of errors made by VLMs and suggest potential pathways for future VLM development. Our dataset and code are available at \url{https://dahlian00.github.io/AgroBenchPage/}.
\end{abstract}    
\begin{table*}[]
\centering
\setlength{\tabcolsep}{4.2pt}
\begin{tabular}{l|c|cc|cc|ccccc}
\toprule
Dataset & Annot. &
Crop & Weed & Disease & Pest &  Images & QA pairs & Main Purpose\\
\midrule

Agri-LLaVA~\cite{agri-llava2024}~&  GPT-4 &  29& - &109 &112 & 391k & 391k (Synthetic) &Training \\
AgroInstruct~\cite{Awais2024AgroGPT} &  GPT-4 & 174 & 4& 74& 12& 70k & 70k (Synthetic) &Training\\
CDDM~\cite{liu2024CDDM} &  GPT-4 &  15 & -- & 60 & -- & 137k & 1M (Synthetic)&Training \\
\midrule
\datasetname (Ours) &  \textbf{Expert} & \textbf{203} & \textbf{108} & \textbf{682} & \textbf{134} & 3,745& 4,342 (\textbf{Expert})&Evaluation\\
\bottomrule
\end{tabular}
\caption{\textbf{Comparison of agricultural vision datasets.} { Expert refers to `human expert' in this context.} \datasetname provides a comprehensive evaluation framework for assessing VLMs, featuring multiple tasks and a wide range of categories.}
\label{tab:dataset_comparison}
\vspace{-11pt}
\end{table*}

\section{Introduction}
\label{sec:intro}

Agriculture is a fundamental process for humans to produce crops to live and stay healthy.
With the development of computer vision technology, effective and automated management of external crop factors such as diseases and pests has been explored, contributing to stable crop production.
This includes detecting and classifying undesirable conditions like diseases~\cite{Singh2020PlantDoc,cacao,sugarcane,potato,IRDD,paddydoc,tomatov} and pests~\cite{agripest,Wu2019Insect, AMRANI2023107587}, as well as general plant management tasks such as crop classification~\cite{Wei2024PlantWild,s23146298} and recognition of crop maturity~\cite{SANYA2024109952,MAITLO2023109462,SHINODA2023100196,LU2022106696}, and structure understanding~\cite{Steininger_2025_WACV,s25144354}.

To maintain stable crop production, it is important to recognize a wide range of crop conditions, including undesirable situations such as diseases and pests, and to know how to respond appropriately. A single visual model that can handle various conditions, not only disease and pest detection but also treatment and general crop management, would be highly beneficial.
However, most existing approaches use task-specific models. These models usually require large amounts of training images and manual annotations for each task. As a result, farmers often need to use several different models depending on the situation. This increases complexity and makes the overall system less accessible for practical use in agriculture.

For general purpose visual tasks, vision-language models (VLMs)~\cite{gpt4o, gpt4omini,gemini,liu2024llavanext,liu2023improvedllava,wang2023cogvlm,Emu} have become widespread recently because they can understand task definitions provided by natural language prompts, covering a wide range of applications without the need for task-specific model training.
The recognition ability of VLMs is closely connected to image recognition itself and linked to an object's words, and supports open-vocabulary recognition through web-scale training. Here, zero-shot recognition and few-shot adaptation have also been realized by VLMs with language representations. This ability opens up a wide range of applications; therefore, we believe this can be applied to agricultural scenarios as well.
It is worth noting that VLMs offer an easy-to-use interface for the general public, especially the question-answer (QA) and conversation modes.
To investigate the range of tasks that VLMs obtained through large-scale vision-language training can cover, recent studies have introduced various benchmark datasets 
to fully evaluate the VLMs, including tasks such as understanding diagrams and charts~\cite{masry-etal-2022-chartqa,Methani_2020_WACV}, and question answering requiring specialized knowledge~\cite{yue2023mmmu,yue2024mmmu,mmstar}.

However, VLM research remains underexplored in agriculture due to a lack of benchmark datasets that include diverse tasks and categories in agriculture. 
Several pioneering works~\cite{liu2024CDDM,agri-llava2024} have adapted open-source VLMs such as LLaVA~\cite{liu2023llava} to the agricultural domain by fine-tuning them on synthetic datasets generated from closed-source VLMs such as GPT-4o.
This is recognized as an effective approach because black-box VLMs often possess a certain amount of agricultural knowledge obtained from the Internet.
While this helps in generating responses that are generally acceptable to experts, there is almost no way to verify whether the answers are truly correct.
Moreover, in the limited evaluation of categories, we are insufficient for fully assessing VLM knowledge; whether they can answer a wide range of types, such as diseases, pests, and weeds.
These limitations motivated us to develop a benchmark dataset in order to evaluate the VLM's broad knowledge in the agricultural domain and its applicability as a practical application.

Here, we introduce \datasetname~(\textbf{Agro}nomist AI \textbf{Bench}mark), a comprehensive benchmark dataset for VLM for the agricultural domain, covering a state-of-the-art range of categories for agriculture-focused benchmark datasets for VLM; 682 disease, 134 pest, 108 weed, and 203 crop categories (Figure~\ref{fig:fig1}).
We cover not only identification tasks but also crop production and disease management knowledge. Moreover, we include other important topics related to crop management with 98 machine categories and 77 traditional management methods to investigate more about VLM's ability.
We carefully selected benchmark tasks from key agricultural engineering research areas, and also the tasks that address challenges faced by farmers in real agricultural scenarios.
We employ a multiple-choice format, and all questions are annotated by human agronomist experts, which overcomes the limitations of previous synthetically created datasets.
We carefully selected images from Creative Commons-licensed and publicly available datasets, including real farm conditions. 
Also, under our manual annotation process, unclear images were removed.
As shown in Table~\ref{tab:dataset_comparison}, \datasetname contains data with the most diverse disease and crop variation categories and many more crop/weed annotations. 

Our main contributions are as follows;
\begin{itemize}
\item We developed \datasetname, a benchmark dataset for VLM, to assess their broad agricultural knowledge of VLMs and evaluate their applicability in practical applications.

\item In our \datasetname evaluation, VLMs tend to achieve high performance in disease and crop management tasks; however, there is still room for improvement in weed and disease identification.

\item Our \datasetname enables error analysis by providing broader category annotations, highlighting future directions for model training focus areas.

\end{itemize}

\begin{figure*}[ht]
\centering
\includegraphics[width=\linewidth]{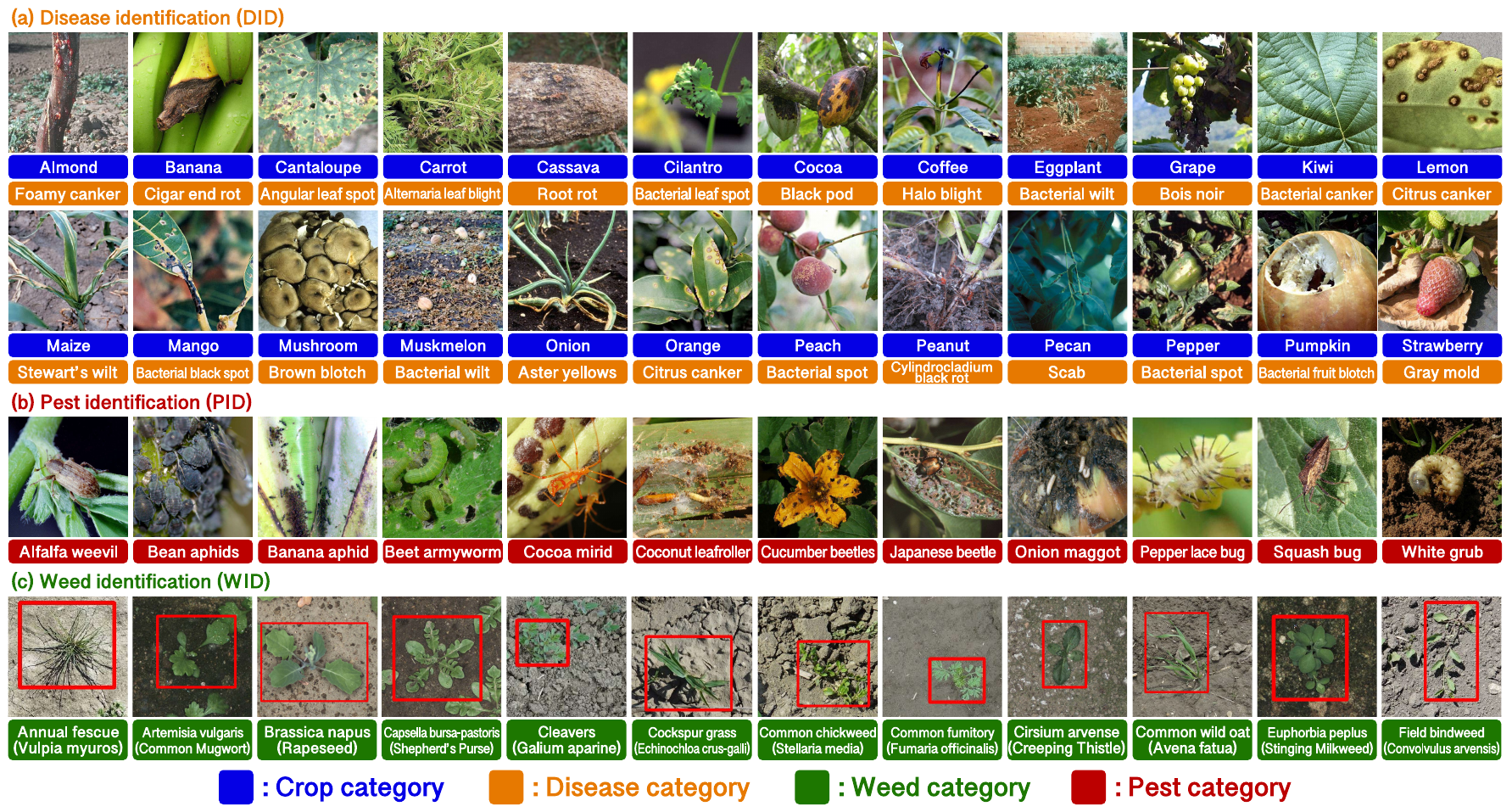}
\caption{\textbf{Examples of labeled images for DID, PID, and WID tasks.}
Our dataset includes 682 crop-disease pairs, 134 pest categories, and 108 weed categories. We prioritized collecting images from real farm settings.}
\vspace{-11pt}
\label{fig:categories}
\end{figure*}

\section{Related Work}
\label{sec:related}

\subsection{Computer Vision for Agriculture}

With the rise of computer vision techniques driven by advancements in deep learning, a wide range of agricultural tasks have been explored over the past decade. In particular, dataset construction and benchmarking play an important role in developing models and practical applications in agriculture, while also paving the way for forward-looking advancements.
Disease identification is a key research focus across various crops, such as rice~\cite{IRDD,paddydoc}, tomato~\cite{tomatov}, cacao~\cite{cacao}, and sugarcane~\cite{sugarcane}.
Multicrop datasets have also been created~\cite{Singh2020PlantDoc, plantv}. PlantDoc~\cite{Singh2020PlantDoc} covers 13 species and 17 classes focused on leaf-based diseases. Plant Village~\cite{plantv} offers 39 classes containing both diseased and healthy leaf categories. 
In addition to crop disease identification, pest identification~\cite{agripest,Wu2019Insect}, weed identification~\cite{haug2014CropWeedFieldCWFID,Olsen2019DeepWeeds,Steininger_2023_WACV} have also been studied. 

While most of the agricultural datasets primarily focus on visual data, few computer vision studies in agriculture investigate the potential of multi-modal approaches.
The PlantWild dataset~\cite{Wei2024PlantWild} includes 56 plant-disease class pairs, which are collected through image search engines. They implement a CLIP-based model and show the possibility of training with combined text and image data.
In an instruction-based format, the CDDM~\cite{liu2024CDDM} has created 16 categories of crops and 60 categories of crop diseases dataset, which generates instructional data using GPT.
While these synthetically created datasets explore the agriculture multi-modal models, there remains a lack of datasets for comprehensive multi-modal model evaluation, validated by human experts and covering a wide range of tasks and categories.

\subsection{Vision-Language Models}
\noindent\textbf{Models.}
Visual models in the computer vision field have been accelerated by language modality with the data resource of sentence-level inputs and web-scale texts. Specifically, CLIP~\cite{clip} has made a significant contribution in this context, by text and image feature alignment through contrastive learning. 
CLIP played a key role in introducing more sophisticated visual representation into language explanations e.g., Flamingo~\cite{AlayracNeurIPS2022_flamingo}, BLIP~\cite{blip-pmlr-v162-li22n,blip2-pmlr-v202-li23q},
Qwen~\cite{qwen}, PaLI~\cite{PaLI}, LLaVA~\cite{liu2023llava,liu2023improvedllava}, CogVLM~\cite{wang2023cogvlm}, and Emu~\cite{Emu,Emu2,wang2024emu3}.
Closed-source VLMs achieve state-of-the-art performance, such as GPT-4o~\cite{gpt4o} and Gemini Pro~\cite{gemini}, which are said to acquire human-level knowledge across diverse fields on the Internet. 

\noindent\textbf{Benchmarks.} Along this line, several recent studies introduced benchmark datasets.
Especially, the representative examples in terms of universal knowledge benchmarking include MMMU~\cite{yue2023mmmu}, MMMU-Pro~\cite{yue2024mmmu} and MMStar~\cite{chen2024we} for multi-modal understanding. These benchmarks contain highly diverse domains (e.g., natural, graph, illustration, medical images) and academic fields (e.g., science, engineering, art, and medical fields) under the tasks of vision language such as question-answering and reasoning. The series of MMMUs have been verified with foundation models like GPT-4V, but they cannot answer the questions perfectly. 
More specific domain datasets have also been proposed, such as those for Medicine~~\cite{royer,matos2024}, Chart~\cite{masry-etal-2022-chartqa,Methani_2020_WACV, shinoda2024sbsfigurespretrainingfigure}, and Video understandings~\cite{LiCVPR2024mvbench,cores2024tvbench,ning2023}, contributing to the accurate evaluation of VLMs and guiding future research directions.

\section{\datasetname~}
\label{sec:dataset}
\begin{figure}[t]
\centering
\includegraphics[width=1.0\linewidth]{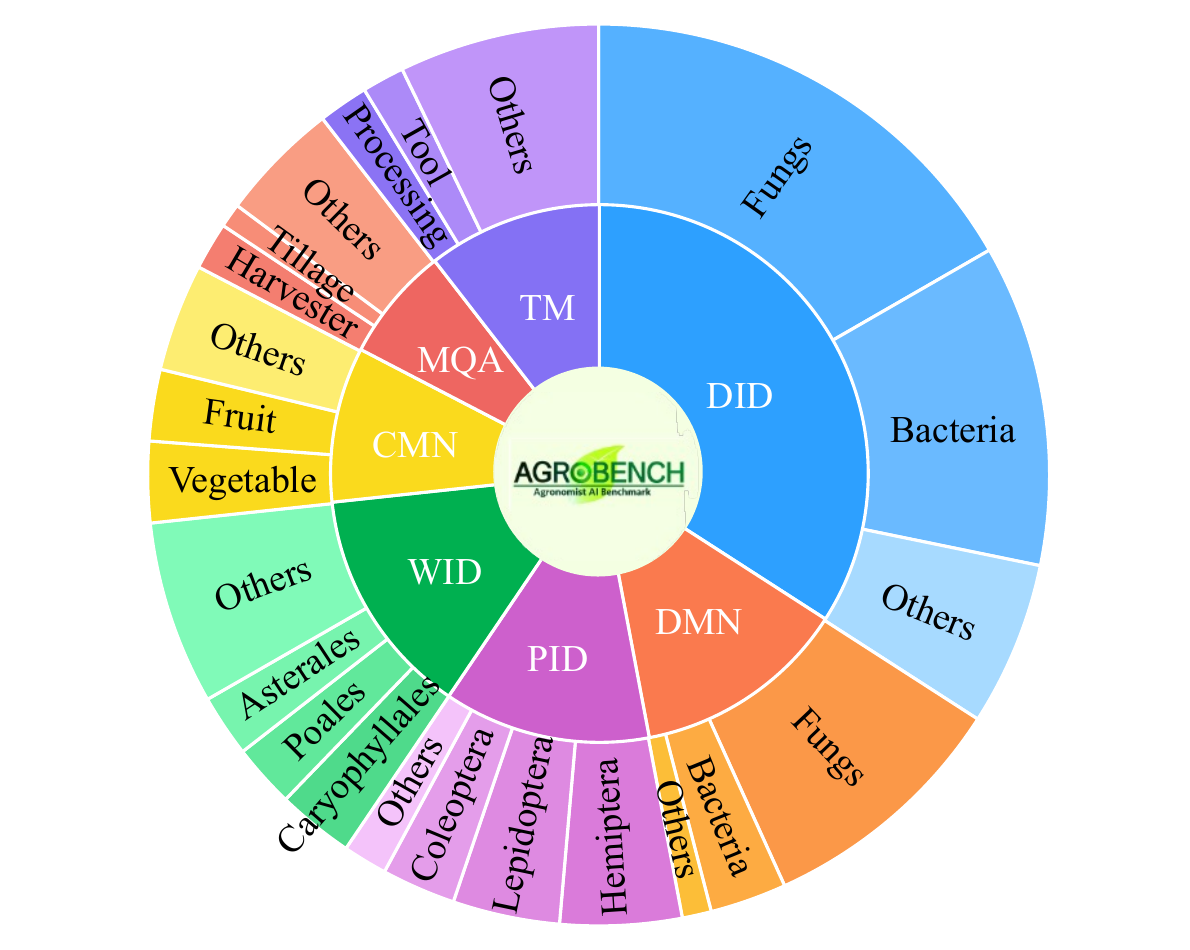}
\caption{\textbf{Seven benchmark tasks in \datasetname.}  \datasetname includes multiple topics with a diverse range of categories. The total accuracy is calculated by the average of each task to mitigate the difference in QAs.}
\label{fig:pie_chart}
\vspace{-15pt}
\end{figure}

\begin{figure*}[ht]
\centering
\includegraphics[width=\linewidth]{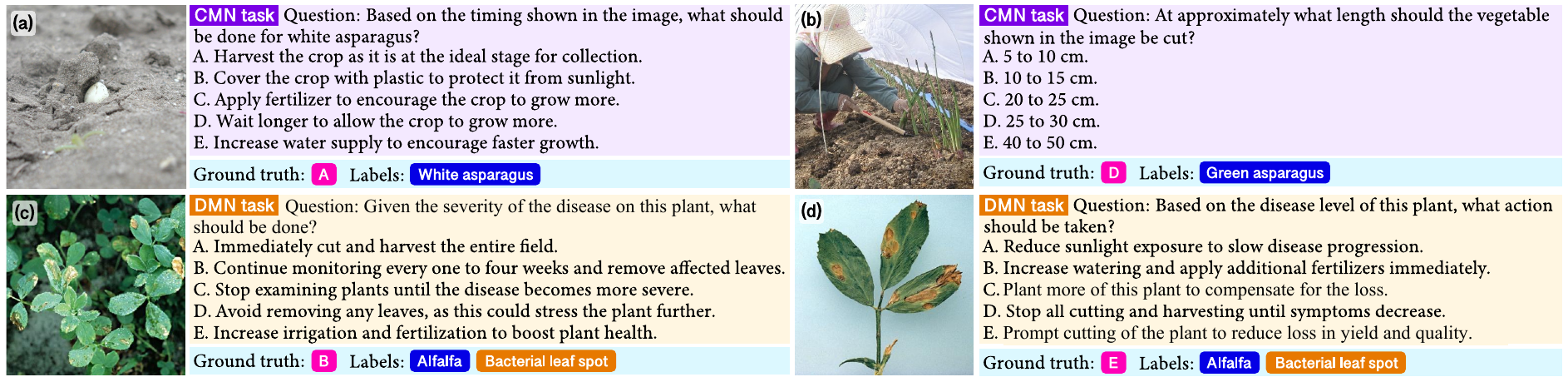}
\vspace{-15pt}
\caption{\textbf{Examples of QA pairs for CMN and DMN tasks.}
(a) and (b) Crop management QA types for
white asparagus and asparagus, respectively.
Their difference in the harvest timing affects the
answer's difference correctly. (c) and (d) Disease management QA types for the alfalfa bacterial leaf spot with the initial and severe symptoms, respectively. Based on the severity of the symptoms, the annotator changes the answer. }
\label{fig:fig3}
\vspace{-11pt}
\end{figure*}

This section introduces \datasetname, the first comprehensive benchmark dataset to evaluate VLM models from the perspective of agricultural vision tasks.
The benchmark consists of seven tasks covering a wide range of tasks selected from key agricultural engineering research areas, as well as tasks that address real-world challenges faced by farmers in real agriculture scenarios. 
\datasetname includes 682 disease categories, 134 pest categories, 203 crop categories, and 108 weed categories, representing the largest number of categories in each area to date, to the best of our knowledge.

\subsection{Benchmark Tasks in Agricultural Scenes}
The seven benchmark tasks encompass key research areas in agricultural engineering as well as real-world challenges faced by farmers.
To facilitate VLM evaluation, we also provide prompts to address each task in a question-answer format. The details of each task are described below.

\noindent\textbf{1) Disease Identification (DID).}
The DID task aims to accurately diagnose and classify crop diseases. This is a key task in agriculture to protect crop health and maximize yields. 
Our benchmark provides 
1,502 QA pairs that cover 370 disease categories, 160 crop categories and 682 crop-disease combinations.
To thoroughly evaluate the capabilities of VLMs, we include four misleading disease labels for each image, featuring diseases with similar symptoms or common diseases affecting the target plant.
VLMs are required to diagnose the disease based on the image, considering both symptoms and the crop species.
Figure~\hyperref[fig:categories]{\ref*{fig:categories}a} illustrates example images.

\noindent\textbf{2) Pest Identification (PID).}
The PID task aims to identify pests to prevent infestations that can severely impact crop health. 
Accurate identification reduces economic losses and minimizes harm to the environment.
Our benchmark provides 544 labeled images that cover 134 pest categories, including insects, mites, and other organisms harmful to plants.
For categories where we obtain multiple images, we select multiple insect growth stages whenever possible. To fully evaluate the VLMs, we assign alternative choices that closely resemble or are commonly associated with the target crops.
Figure~\hyperref[fig:categories]{\ref*{fig:categories}b} shows examples of labeled pest images.

\noindent\textbf{3) Weed Identification (WID).}
The WID task aims to identify weed species.
Our benchmark includes 609 images of weeds with ground-truth bounding boxes, covering 108 weed species commonly found in farm fields.
We assign bounding boxes because multiple weed types often grow closely together, and we want to clarify which one is the target.
Specifically, VLMs are required to identify the weed species within the provided bounding box on the image.
Figure~\hyperref[fig:categories]{\ref*{fig:categories}c} shows examples of labeled weed images.

\noindent\textbf{4) Crop Management (CMN).}
Crop management focuses on optimizing farming practices to facilitate crop growth.
This involves making decisions on irrigation, fertilization, planting times, and other cultivation practices.
Our benchmark provides 411 question-answer pairs for this task.
VLMs are required to analyze images of crops and recommend appropriate management strategies by considering factors such as crop health, growth stage, and environmental conditions visible in the images given five answer candidates.
In Figure~\hyperref[fig:fig3]{\ref*{fig:fig3}a} and~\hyperref[fig:fig3]{b}, we show two examples of the harvest timing for white and green asparagus, respectively. Our dataset includes complex questions that take into account the differences in their harvest timing. 

\noindent\textbf{5) Disease Management (DMN).}
Disease management aims to control and reduce diseases in crops. This involves informed decisions on interventions such as applying pesticides, adopting resistant crop varieties, or modifying cultivation practices.
Our benchmark provides 569 question-answer pairs covering 141 crop-disease combination categories. Since many diseases share the same management strategies regardless of the crop, we carefully select a diverse range of disease types and management strategies.
VLMs are required not only to identify the disease but also to recommend appropriate management strategies based on images of affected crops.
In Figure~\hyperref[fig:fig3]{\ref*{fig:fig3}c} and~\hyperref[fig:fig3]{d} show example QAs from the Disease Management tasks. Both images depict Bacterial leaf spot on alfalfa, with c showing the initial stage of the disease and d showing the severe stage. Based on the severity of the disease, we provide different answer options and set different correct answers.
The input prompt consists of a question and five answer candidates.

\noindent\textbf{6) Machine Usage QA (MQA).}
Machine usage QA addresses the correct use and choice of agricultural machinery depending on the task and farming conditions.
Selecting the appropriate machinery is essential for efficient farming.
Our benchmark provides 303 question-answer pairs covering 98 machine categories. Given that many crops share the same machinery (e.g., soil preparation and irrigation), the coverage of this category is comprehensive.
VLMs are required to answer questions about machinery operation or select the appropriate machine for a given scenario based on images of machinery or farming conditions.

\noindent\textbf{7) Traditional Management (TM).}
Traditional management methods involve natural and sustainable approaches to farming, such as the use of organic fertilizers, terrace farming, and agroforestry. Recent computer vision studies have not focused on these traditional practices, although many are still used by certain local farmers. Our benchmark includes 404 question-answer pairs, including 77 traditional management practices. VLMs are required to identify the management method or explain its effectiveness, given five answer choices.

\subsection{Dataset Construction}

\begin{table*}[h]
\centering
\begin{tabular}{l|cccccccccc}
\toprule
Model& DID & DMN & PID & WID& CMN&MQA&TM&Overall &Overall \\
& & && & &&&(all)&(subset)\\
\midrule
Random Choice& 21.77 & 15.64 & 20.40 & 17.90 & 16.06 & 22.11 & 19.31 & 19.03 &19.11\\[-0.5ex]
Human&25.00  &22.50 &45.00 & 20.00 &36.25 &57.50&51.25 &-&36.79\\[-0.5ex]
\midrule
\multicolumn{10}{c}{\textbf{Closed-Source Vision Language Models (VLMs)}} \\ \midrule
GPT-4o mini~\cite{gpt4omini} & 53.60 & 80.67 & 60.04 & 35.14 & 64.23 & 70.96 & 69.80 & 62.06 & 69.65 \\
GPT-4o~\cite{gpt4o} &  64.18 & 87.35 & 77.76 & 44.17 & 75.43 & 82.84 & 82.43 & 73.45 & 79.26 \\
Gemini1.5-Flash~\cite{gemini}  &55.06 & 79.96 & 70.04 & 50.90 & 64.72 & 78.22 & 73.27 & 67.45 & 68.82 \\
Gemini1.5-Pro~\cite{gemini}  &62.92 & 81.55 & 74.45 & 55.17 & 71.05 & 82.84 & 77.72 & 72.24 & 69.74 \\\midrule
\multicolumn{10}{c}{\textbf{Open-Source Vision Language Models (VLMs)}} \\ \midrule
EMU2Chat~\cite{Emu2} & 42.01 & 48.33 & 43.75 & 23.81 & 40.39 & 37.62 & 47.77 & 40.53 & 33.84 \\
LLaVA-Next-8B~\cite{liu2024llavanext}&45.47 & 72.58 & 43.01 & 30.05 & 54.26 & 56.11 & 57.46 & 51.28 & 57.84 \\
LLaVA-Next-72B~\cite{liu2024llavanext}& 54.95 & 80.00 & 49.81 & 26.98 & 66.92 & 66.11 & 70.38 & 59.31 & 64.36 \\
QwenVLM-7B~\cite{qwen} & 51.26 & 80.49 & 63.97 & 33.17 & 66.42 & 76.24 & 77.48 & 64.15 & 66.41 \\
QwenVLM-72B~\cite{qwen} &57.99 & 87.87 & 73.35 & 34.48 & 75.91 & 80.86 & 84.16 & 70.66 & 72.45 \\
CogVLM-19B~\cite{wang2023cogvlm} &29.16 & 53.78 & 52.39 & 25.45 & 54.01 & 71.62 & 66.09 & 50.36 & 44.27 \\
LLaVa-7B~\cite{liu2023llava} & 36.02 & 62.74 & 38.79 & 24.79 & 53.77 & 46.53 & 55.20 & 45.41 & 46.14 \\
LLaVA-13B~\cite{liu2023llava} & 40.21 & 68.89 & 44.49 & 24.79 & 59.37 & 54.13 & 58.42 & 50.04 & 55.31 \\
\bottomrule
\end{tabular}
\caption{\textbf{Results for the seven benchmark tasks with images.} We provide results for Random Choice, Human Validation, four closed-source VLMs, and open-source VLMs. Human validation was conducted by 28 people on a subset of 80 samples per task as a reference.}
\label{tab:baseline}
\end{table*}
\begin{table*}[t]
\centering
\begin{tabular}{l|cccccccccc}
\toprule
\multirow{2}{*}& DID & DMN & PID & WID & CMN & MQA & TM & Overall \\
[-0.5ex] \midrule
GPT-4o~\cite{gpt4o} & 1.93 & 72.58 & 18.75 & 1.00 & 40.39 & 25.08 & 48.27 & 29.71  \\
LLaVA-Next-8B~\cite{liu2024llavanext} & 26.10 & 70.30 & 21.88 & 19.70 & 53.77 & 30.36 & 40.35 & 37.49  \\

\bottomrule
\end{tabular}
\caption{\textbf{Results for the seven benchmark tasks with text only.} We additionally evaluate the model without image inputs, and the overall performance is close to random.}
\vspace{-15pt}
\label{tab:textinput}
\end{table*}

\noindent\textbf{Image Selection.}
To establish a high-quality benchmark covering a wide range of crops, diseases, and pest categories we initially curated around 50,000 agricultural images from websites supervised by plant pathologists, either where redistribution was permitted or where we obtained redistribution permission.
We selected images in real farm settings as much as possible, as shown in Figure~\ref{fig:categories}, to evaluate real-world scenarios. When licensed images for a target category were limited, we used laboratory setting images.
For curation, we obtained images along with their corresponding labels. 
The annotator is one of the authors, who holds a Ph.D. in Agriculture.
In the human evaluation conducted during the experiment, other individuals with a Ph.D. or M.S. degree in Agriculture reviewed questions to assess the quality of the QA pairs.
The annotator selected images that clearly represented target labels and removed those images with irrelevant labels.
For the Machine usage QA and Traditional management, annotators manually created categories based on textbooks and websites, and searched for corresponding images with redistribution licenses.
For Weed identification, we use existing dataset~\cite{Steininger_2023_WACV, OPPD,cottonweed, Perrenial}. For this Weed identification category, we will provide a simple code to download the data and crop images and assign bounding boxes for Weed Identification, allowing users to work with the existing dataset without redistributing the images.
Through these image selection processes, the selected images are high-quality 4,218 representative images.
The detailed distribution of tasks and categories for each task is summarized in Figure~\ref{fig:pie_chart}.

\noindent \textbf{QA Annotation.}
For the DMN, CMN, MQA, and TM tasks, all question-answer pairs were created manually, independent of any LLMs or VLMs.
Annotators used GPT only for sentence rephrasing, but they were prohibited from using any knowledge from GPT for QA creation.
Annotators were allowed to refer to textbooks, academic journals, and other authoritative sources in the field to ensure the accuracy and depth of the dataset. 
This took around 150 man-hours. We carefully annotate various types of questions. 
Examples of QAs that require expert knowledge are shown in Figure~\ref{fig:fig3}.
All questions were carefully created so that image reference is necessary for answering.

\noindent \textbf{Dataset Statistics.} 
Following the dataset annotations and selections, our \datasetname comprises seven tasks with 4,342 QA pairs, as shown in Table~\ref{tab:dataset_comparison}.
All tasks comprise a wide range of categories for detailed evaluation. Please refer to the supplementary materials for more details and examples.
\begin{figure*}[t]
\centering
\includegraphics[width=\linewidth]{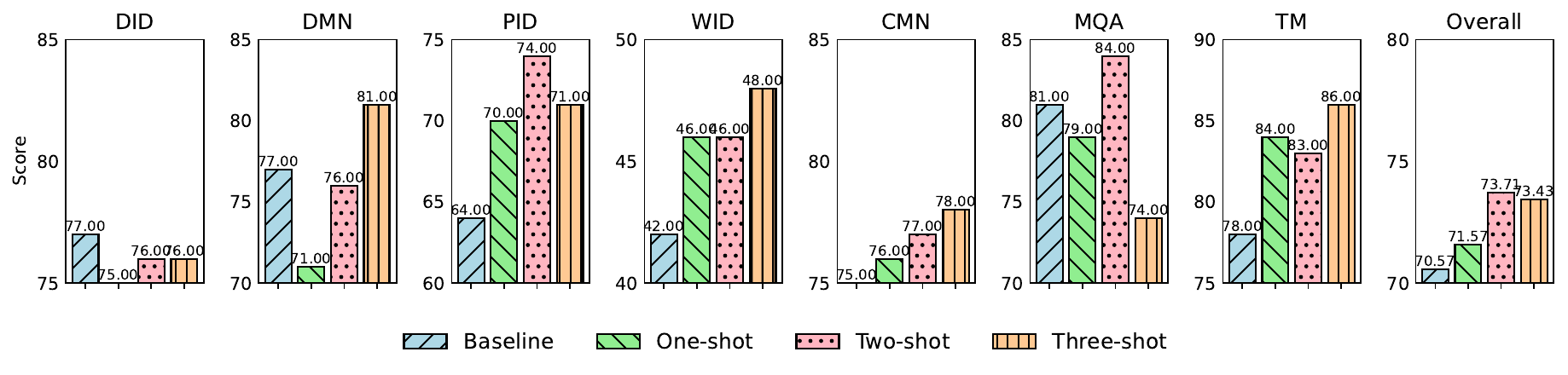}
\vspace{-15pt}
 \caption{ \textbf{Results of seven benchmark tasks with Chain of Thought~(CoT).} Baseline indicates results without CoT.  
In the one-shot, two-shot, and three-shot settings, we provide one, two, and three CoT examples per task, respectively, to guide the model.}
\label{fig:fig4}
\end{figure*}
\begin{figure*}[ht]
\centering
\includegraphics[width=\linewidth]{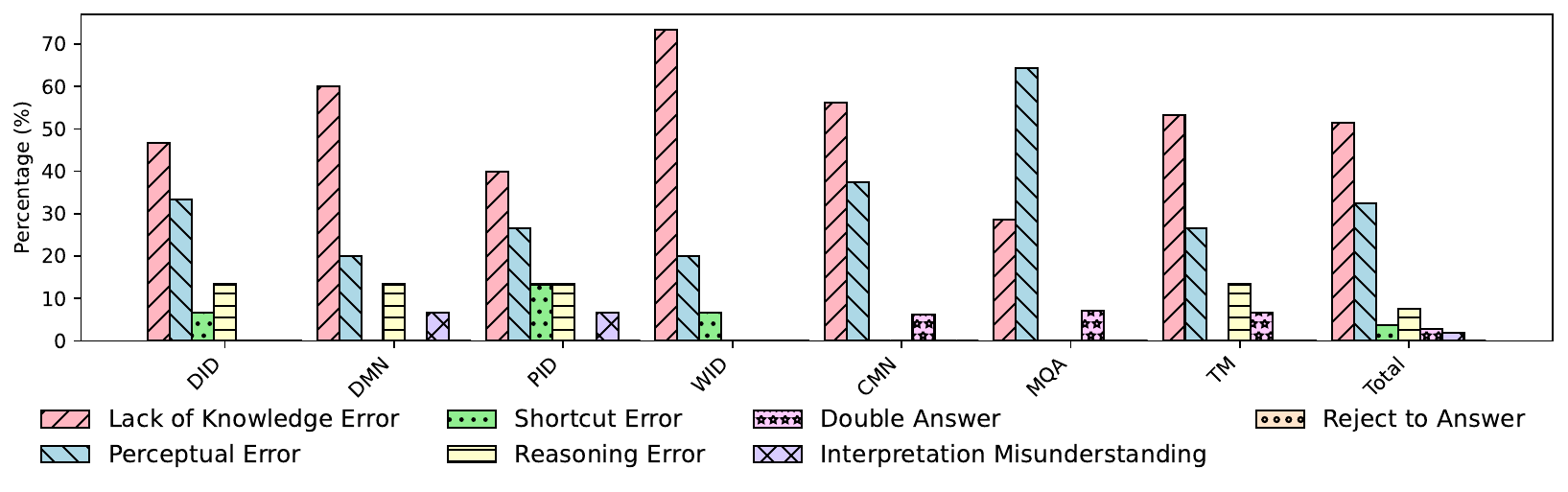}
\vspace{-15pt}
 \caption{ \textbf{Error analysis on seven benchmark tasks with GPT-4o.} We extract a maximum of 15 errors per task from the zero-shot CoT result. We manually analyze how they conclude the incorrect answer.}
 \vspace{-15pt}
\label{fig:fig5}
\end{figure*}

\section{Experiments}
\subsection{Experimental Settings}
\noindent\textbf{Baseline Models.}
We use four closed-source models: GPT-4o~\cite{gpt4o}, GPT-4o mini~\cite{gpt4omini}, Gemini1.5-Pro~\cite{gemini}, and Gemini1.5-Flash~\cite{gemini}.
GPT-4o mini and Gemini1.5-Flash are down-scale versions of GPT-4o and Gemini1.5-Pro, respectively.
We use eight open-source models: EMU2Chat~\cite{Emu2}, LLaVA-Next-8B~\cite{liu2024llavanext}, LLaVA-Next-72B~\cite{liu2024llavanext}, QwenVLM-7B~\cite{qwen}, QwenVLM-72B~\cite{qwen}, CogVLM-19B~\cite{wang2023cogvlm}, LLaVa-7B~\cite{liu2023llava}, and LLaVa-13B~\cite{liu2023llava}. For the details of these models, please refer to the supplementary material.

\noindent\textbf{Human Results.}
We also present results from human participants for reference. We surveyed 28 students, each holding at least a bachelor's degree in agriculture, and asked them to answer 20 questions each. This created a test subset of 280 questions, with each question answered by two participants, resulting in a total of 560 responses.
We averaged the results per task and reported the accuracy.
Each participant was permitted to use a book or translator to look up word meanings but was prohibited from using the internet for searches. If participants were unsure of the answer, they were asked to provide the response they believed to be most accurate.

\noindent\textbf{Evaluation Protocol.}
Importantly, our dataset evaluation is conducted per task, and overall scores are averaged based on the number of tasks, not the number of QAs. This prevents categories with a large number of evaluations from becoming dominant.
We adopt an exact matching approach for our five-option questions. If the model's response matches the option’s letter or the answer sentence, we consider it correct. 
If the model’s answer does not match any option, including cases where there is no answer or multiple answers, we consider it incorrect.

\subsection{Main Results}
Here, we discuss the main results of our \datasetname evaluation performance. We evaluate our \datasetname using eight open-source VLMs and four closed-source VLMs with APIs.

\noindent\textbf{{Challenges of the \datasetname.}} We show the main results in Table~\ref{tab:baseline}.
The most difficult task is Weed Identification (WID), on which most open-source VLMs perform at around the random score. The highest WID accuracy, 55.17\%, was achieved by Gemini 1.5-Pro.
This suggests that the knowledge about weeds is not as fully trained as that of crops.
All models' Disease Identification (DID) results are lower than their Disease Management (DMN) results. 
This means that VLM models can gain contextual information, but there is still room for perceptual improvement.

\noindent\textbf{{Model Comparison.}}~Overall, closed-source VLMs achieve better results than open-source VLMs and achieve higher performance than humans. GPT-4o model achieved the highest score in overall performance.
Among the open-source models, QwenVLM-72B achieves the best result in overall accuracy, which is comparable or even superior to open-source VLMs in some tasks. 
QwenVLM-72B achieves satisfactory results on both identification and question-answering benchmarks on \datasetname.

\subsection{Ablations}
\noindent\textbf{{Context Reliance.}}~We further evaluate the results on \datasetname using text-only input to determine whether visual information is necessary to answer the questions. Table~\ref{tab:textinput} presents the experimental results across seven benchmark tasks on \datasetname.
With text-only input, performance drops across all models, confirming the reliance on visual information. However, for both models, the DMN, CMN, and TM tasks maintain significantly higher accuracy than random selection. Although we ensured that our questions do not include disease names or appearance-related traits, models tend to infer answers based on estimation. This suggests that many disease types share common management strategies, such as avoiding humidity, preferring cooler temperatures, or pruning infected parts, allowing models to predict the most likely option.
A similar pattern is observed in the CMN and TM tasks, where models can make educated guesses based on contextual cues. (See the supplementary material for further detailed examples.)

\noindent\textbf{{Chain of Thought.}} 
We evaluate the effectiveness of Chain of Thought (CoT) reasoning. 
In the one-shot setting, we provide a single CoT reasoning example along with an image and the corresponding prompt for each task to guide the model in answering the question. We select 100 QAs per task for evaluation and use GPT-4 as the VLM.
Figure~\ref{fig:fig4} presents the accuracy of CoT reasoning with the baseline results.
While CoT achieves slightly higher accuracy, the improvement is not significant compared to the baseline. However, CoT demonstrates effectiveness in certain tasks, such as PID, WID, CMN, and TM.
For instance, in PID, the VLM performs step-by-step and careful reasoning (see the supplementary material for more detailed examples). In WID, the most challenging task, CoT provides useful examples that may help the model make predictions.
Overall, CoT contributes to accuracy, but we observe performance saturation in the three-shot setting.
For further detailed examples, please refer to the supplementary material.

\subsection{Error Analysis}
\begin{figure}[t]
\centering
\includegraphics[width=\linewidth]{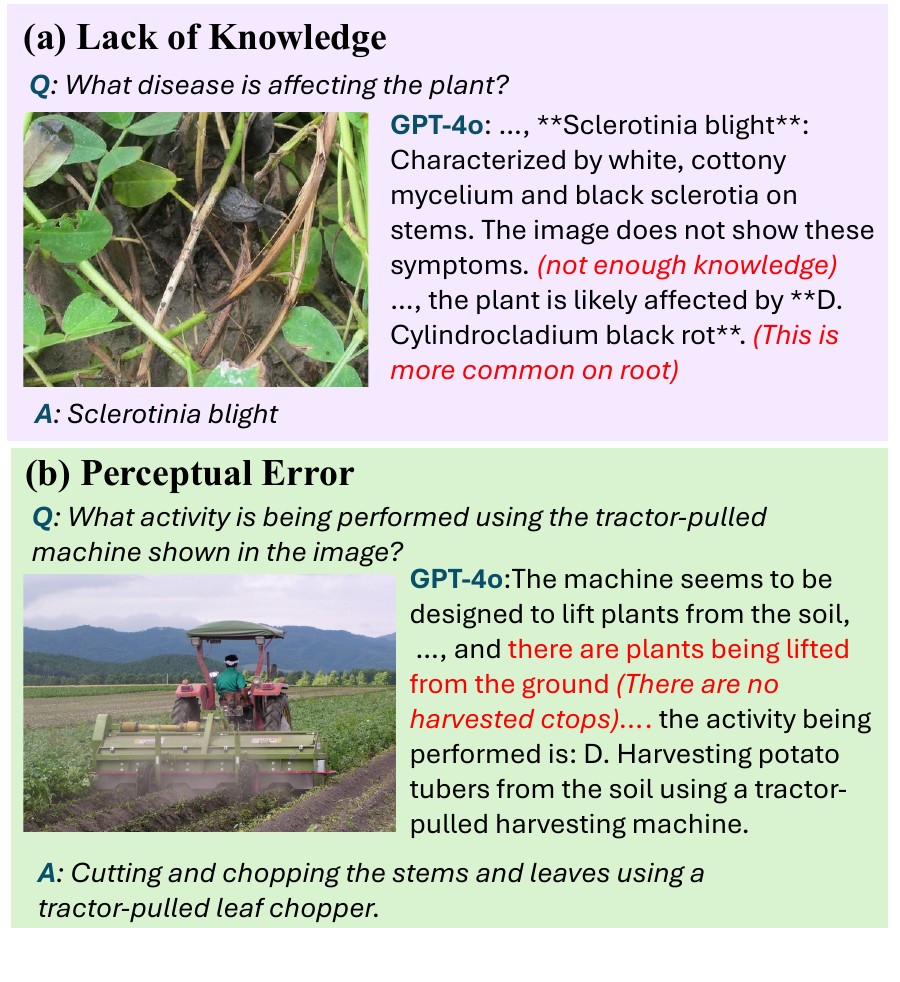}
\vspace{-30pt}
 \caption{ \textbf{Error examples of GPT-4o.} Examples illustrate the two main error types: Lack of Knowledge and Perceptual Error.}
\label{fig:fig6}
\vspace{-15pt}
\end{figure}

Here, we analyze the mistake types that occur depending on the agricultural tasks. We bring out up to 15 failure examples per task from the zero-shot CoT results and manually analyze how they reached the incorrect answers. 

\noindent\textbf{Lack of Knowledge~(51.92\%).}~This includes cases where VLM can’t accurately describe the appearance or relevant knowledge of a choice (e.g., VLM fails to describe disease symptoms or insect characteristics) or lacks context (e.g., VLM doesn’t know how to treat diseased crops or manage crops for high yield).
Figure~\hyperref[fig:fig6]{\ref*{fig:fig6}a} shows an example of a Lack of Knowledge case. When the VLM analyzes the correct answer option for Sclerotinia blight, it fails to describe the symptoms of discoloration and wilting. Additionally, its incorrect answer choice is more commonly associated with the root rather than the stem.
These errors are based on a Lack of Knowledge, suggesting VLMs need more detailed categories and domain-specific training.

\noindent\textbf{Perceptual Error~(32.69\%).}~ 
This indicates that VLM can't pay attention or recognize the answer-related part in the image~(e.g., can't recognize the green insect on the leaf), and VLM misunderstands the image, leading to incorrect answers. Figure~\hyperref[fig:fig6]{\ref*{fig:fig6}b} shows an example of the model hallucinating and misunderstanding the situation. First, the VLM incorrectly identifies the machine as one used for lifting plants from the soil. Then, it describes the scene as if there are harvested crops, even though no harvested crops are present.
These errors can be mitigated by enhancing the VLM’s perception abilities for domain-specific classes. Additionally, improving general perception capabilities, including reducing hallucinations, can further contribute to VLM performance.

\noindent\textbf{Reasoning Error~(7.6\%).}~Reasoning error involves the VLM can describe the options correctly, but can't compare them step by step and conclude the wrong answer.
This error is relatively low compared to the existing work~\cite{yue2023mmmu} since \datasetname requires more specific knowledge and doesn't include reasoning relying on problems (e.g., math).

\noindent\textbf{Other Errors~(7.79\%).}~ For the other errors, we observe Shortcut Error (The VLM can pick up the two candidate options correctly but conclude the answer without comparing the candidates), Double Answer Error (Concluding two answers are correct), Interpretation Misunderstanding~(VLM misleading the question and conclude wrong answer), and Reject to Answer~(VLM conclude there is no answer). 

\section{Conclusion}
In this paper, we develop \datasetname, a comprehensive benchmark dataset for VLMs in the agricultural domain, covering a state-of-the-art range of categories.
Our dataset comprises seven benchmark tasks encompassing key research areas in agricultural engineering as well as real-world challenges faced by farmers.
\datasetname contributes to agricultural VLM research by addressing the lack of datasets for comprehensive multi-modal model evaluation, validated by human experts.

In our evaluation, VLMs exhibit strengths across different tasks. However, in several tasks such as weed identification and disease identification, all models show room for improvement. Our error analysis reveals that most failures are due to a lack of knowledge (51.92\%), suggesting that VLMs require more specialized agricultural knowledge. Our dataset will facilitate agricultural VLM research, enabling broad category and task evaluation to support sustainable, automated agriculture.

\section{Acknowledgment}
This work was supported by the AIST KAKUSEI Project (FY2024) and JST FOREST Grant Number JPMJFR206F. We would like to thank the Agricultural Administration Division, Department of Agriculture, Hokkaido Government, for providing some of the images. We are also grateful to Daniel Steininger for contributing images from the CropAndWeed dataset to our dataset.
We used ABCI 3.0 provided by AIST and AIST Solutions.
{
    \small
    \bibliographystyle{ieeenat_fullname}
    \bibliography{main}
}

\clearpage
\appendix 

\section{Statistics}
\label{sec:statistics}
\begin{figure*}[t]
    \centering
    \includegraphics[height=0.9\textheight]{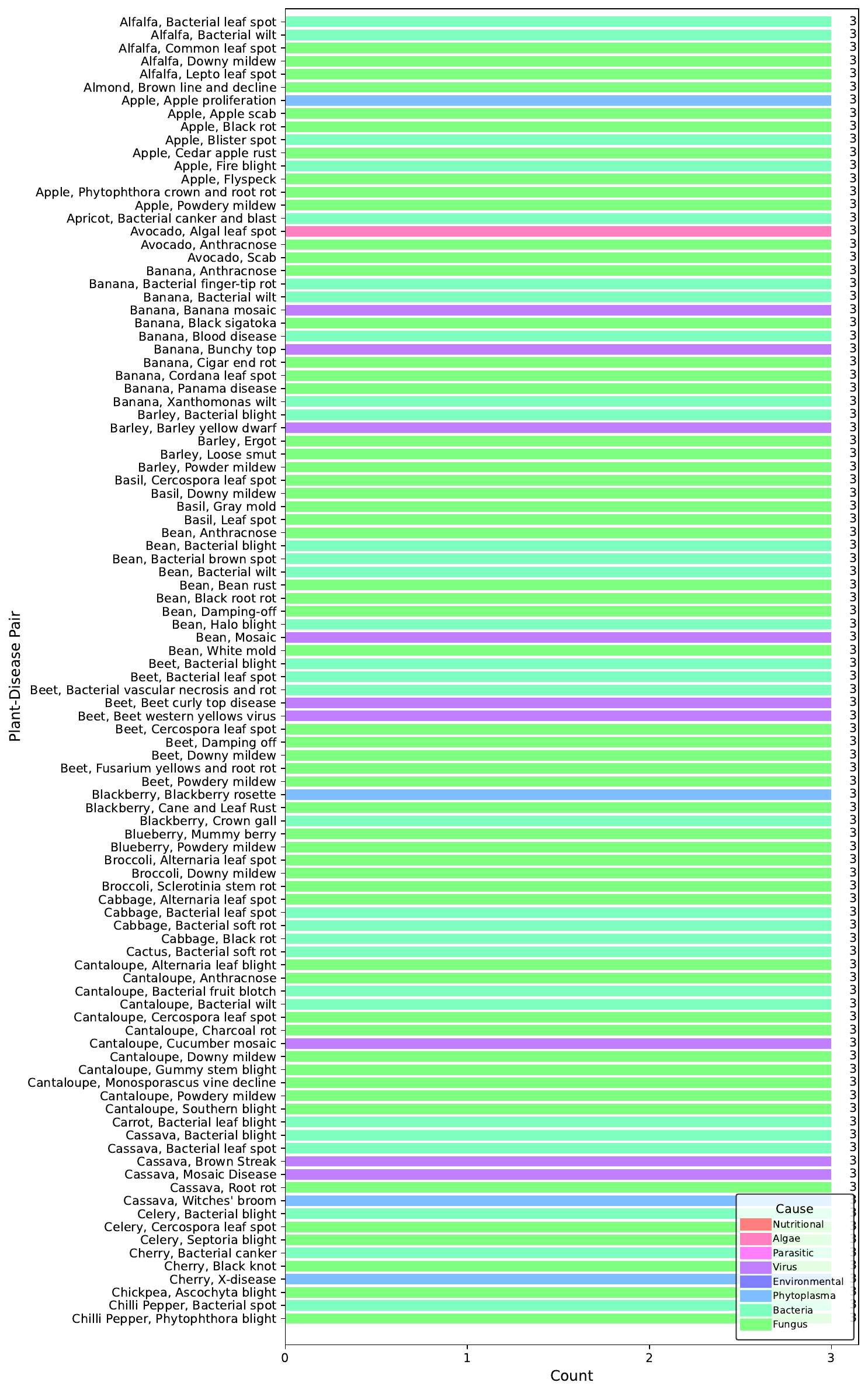}
  \caption{\textbf{Disease Categories Counts (A).}}
\label{fig:diseasecountA}
\vspace{-11pt}  
\end{figure*}

\begin{figure*}[t]
    \centering
    \includegraphics[height=0.9\textheight]{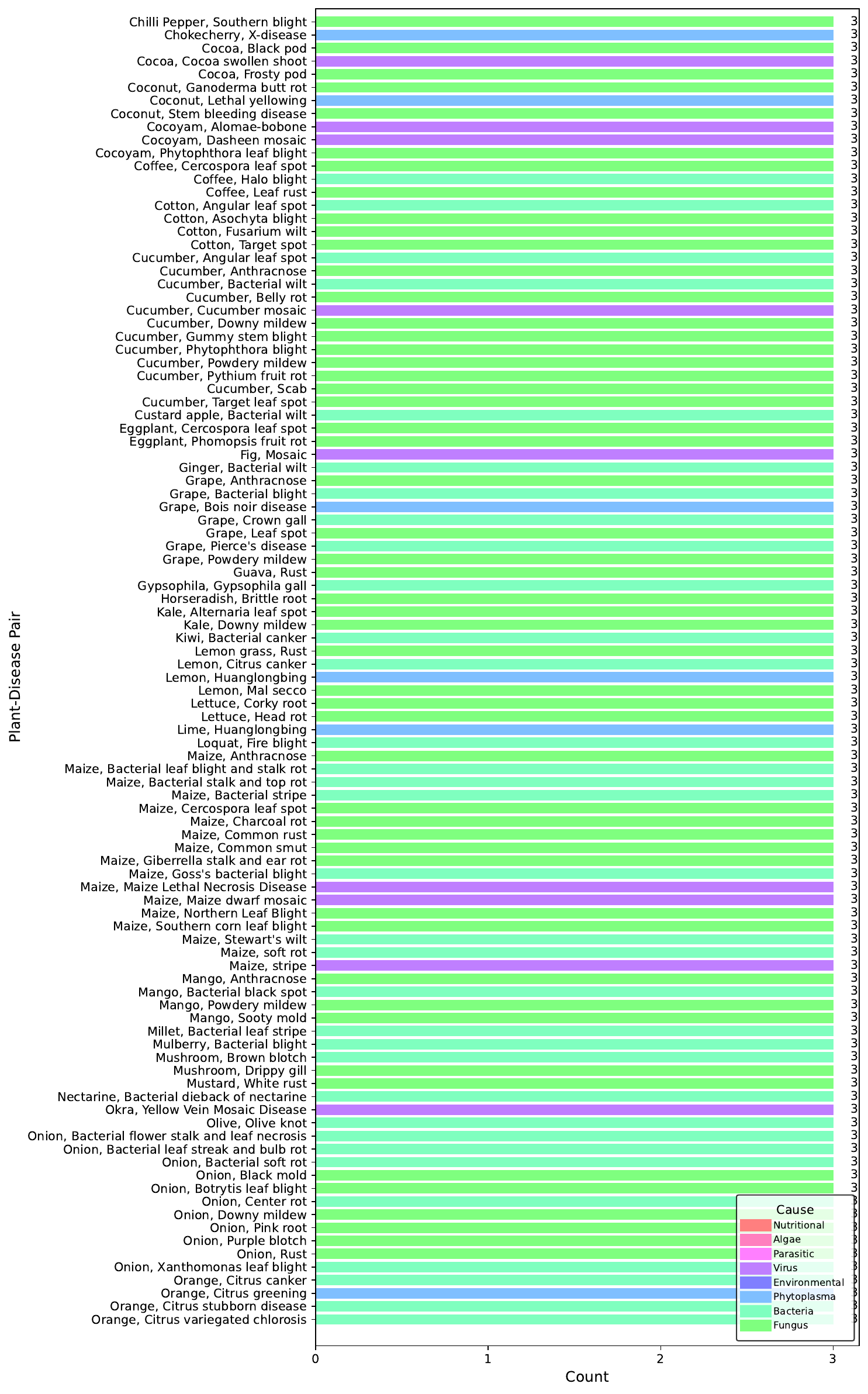}
\caption{\textbf{Disease Categories Counts (B).}}
\label{fig:diseasecountB}
\vspace{-11pt}
\end{figure*}

\begin{figure*}[t]
\centering
    \includegraphics[height=0.9\textheight]{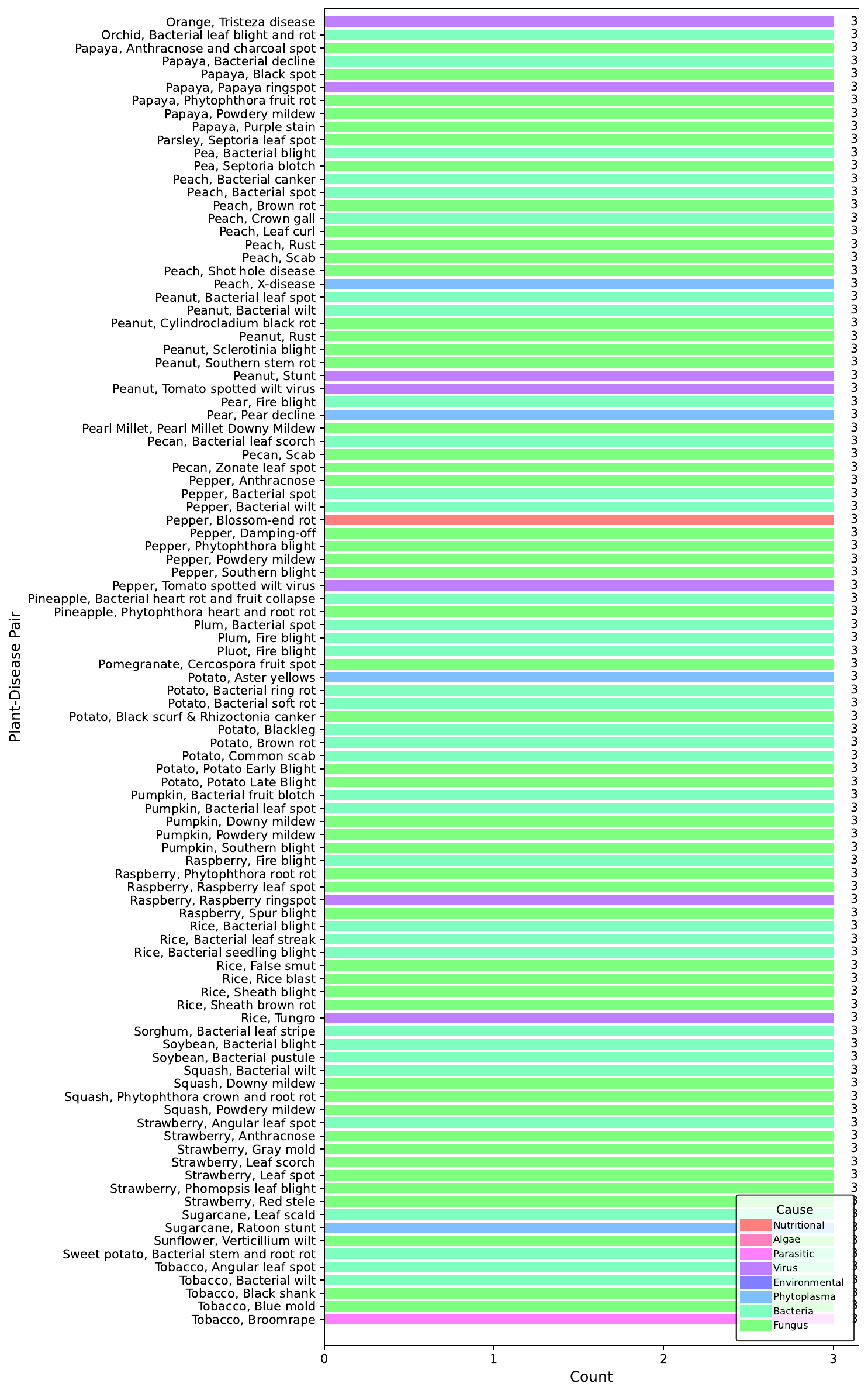}
\caption{\textbf{Disease Categories Counts (C).}}
\label{fig:diseasecountC}
\end{figure*}

\begin{figure*}[t]
    \centering
    \includegraphics[height=0.9\textheight]{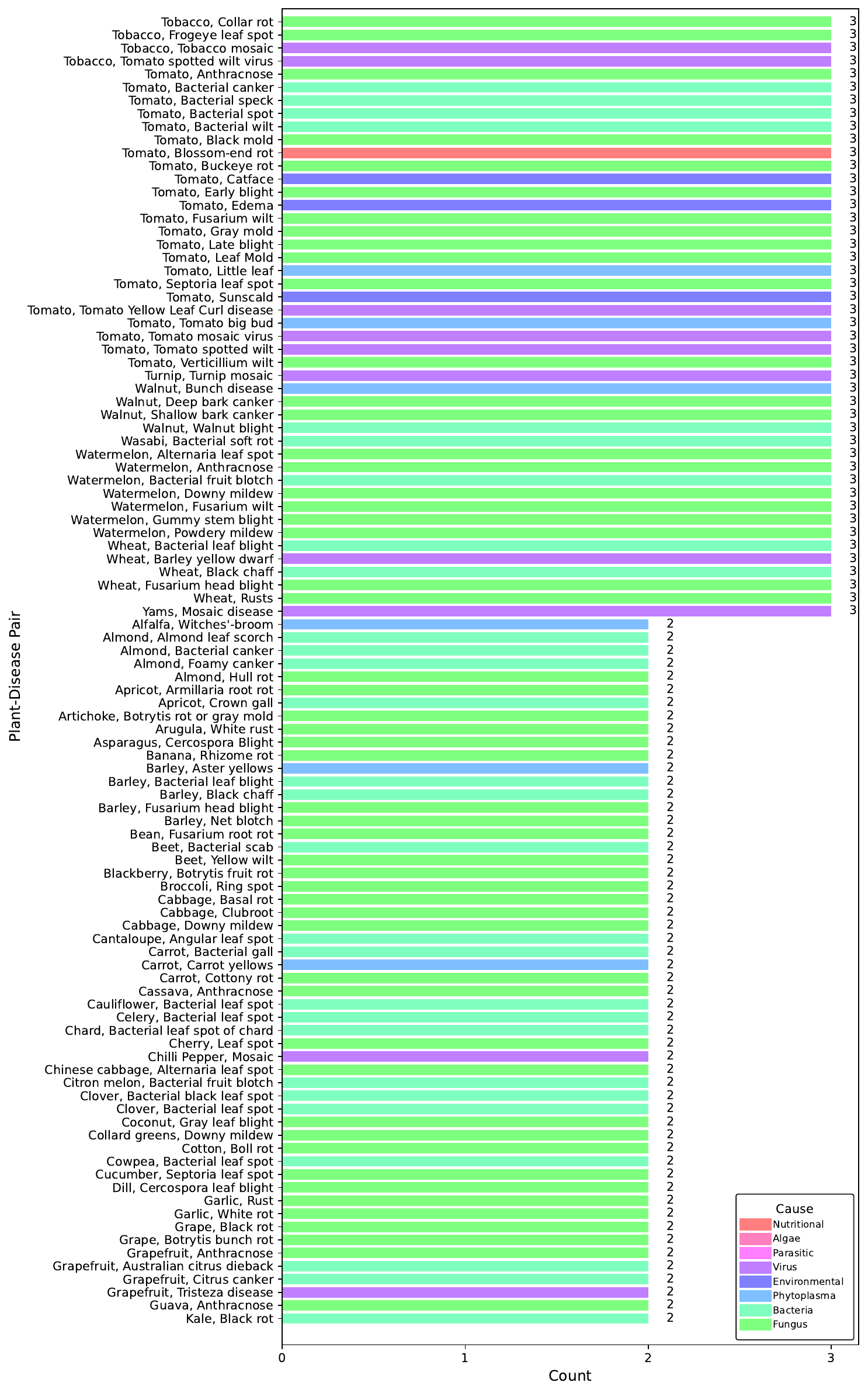}
\caption{\textbf{Disease Categories Counts (D).}}
\label{fig:diseasecountD}
\vspace{-11pt}
\end{figure*}

\begin{figure*}[t]
\centering
    \includegraphics[height=0.9\textheight]{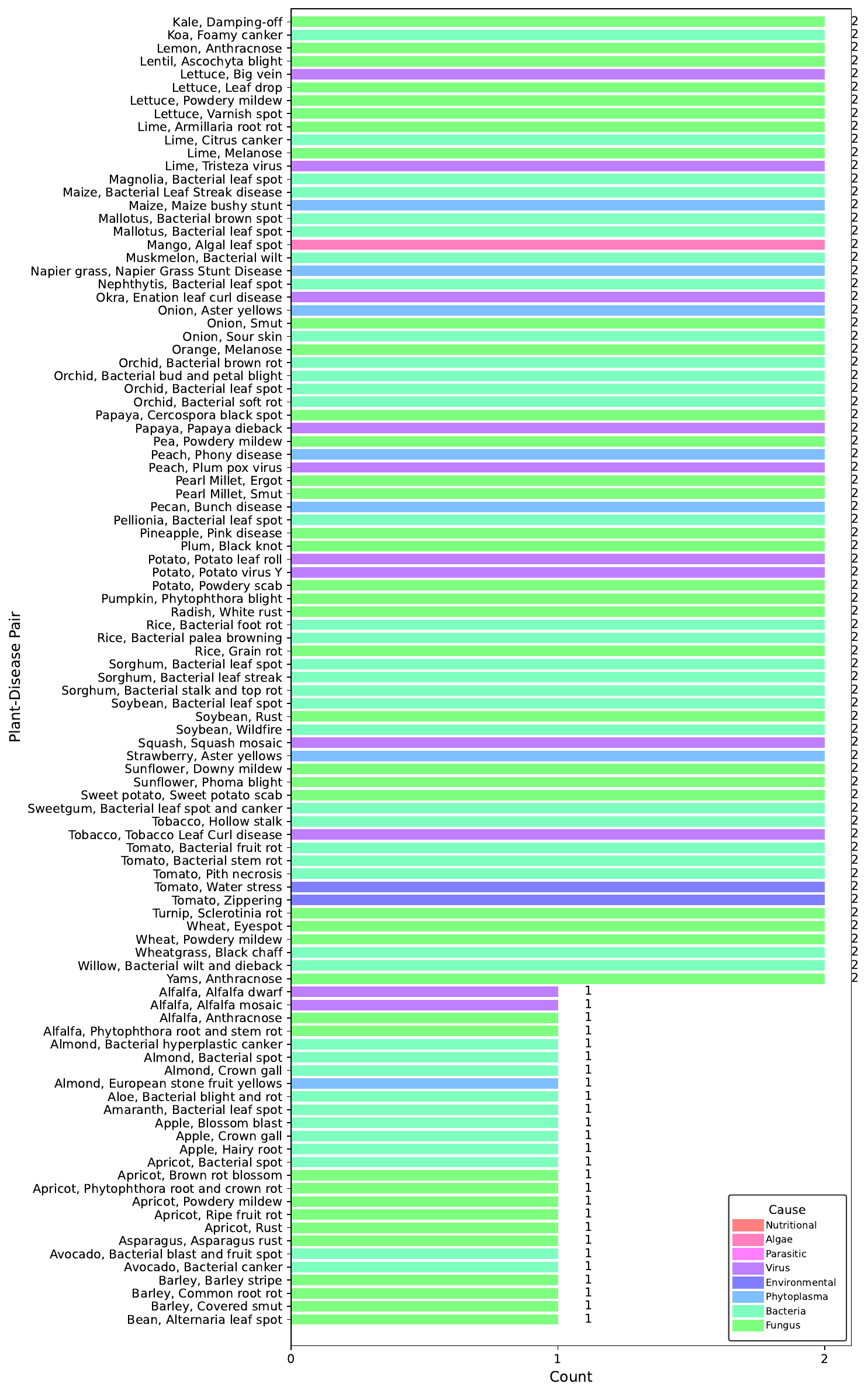}
\caption{\textbf{Disease Categories Counts (E).}}
\label{fig:diseasecountE}
\end{figure*}

\begin{figure*}[t]
    \centering
    \includegraphics[height=0.9\textheight]{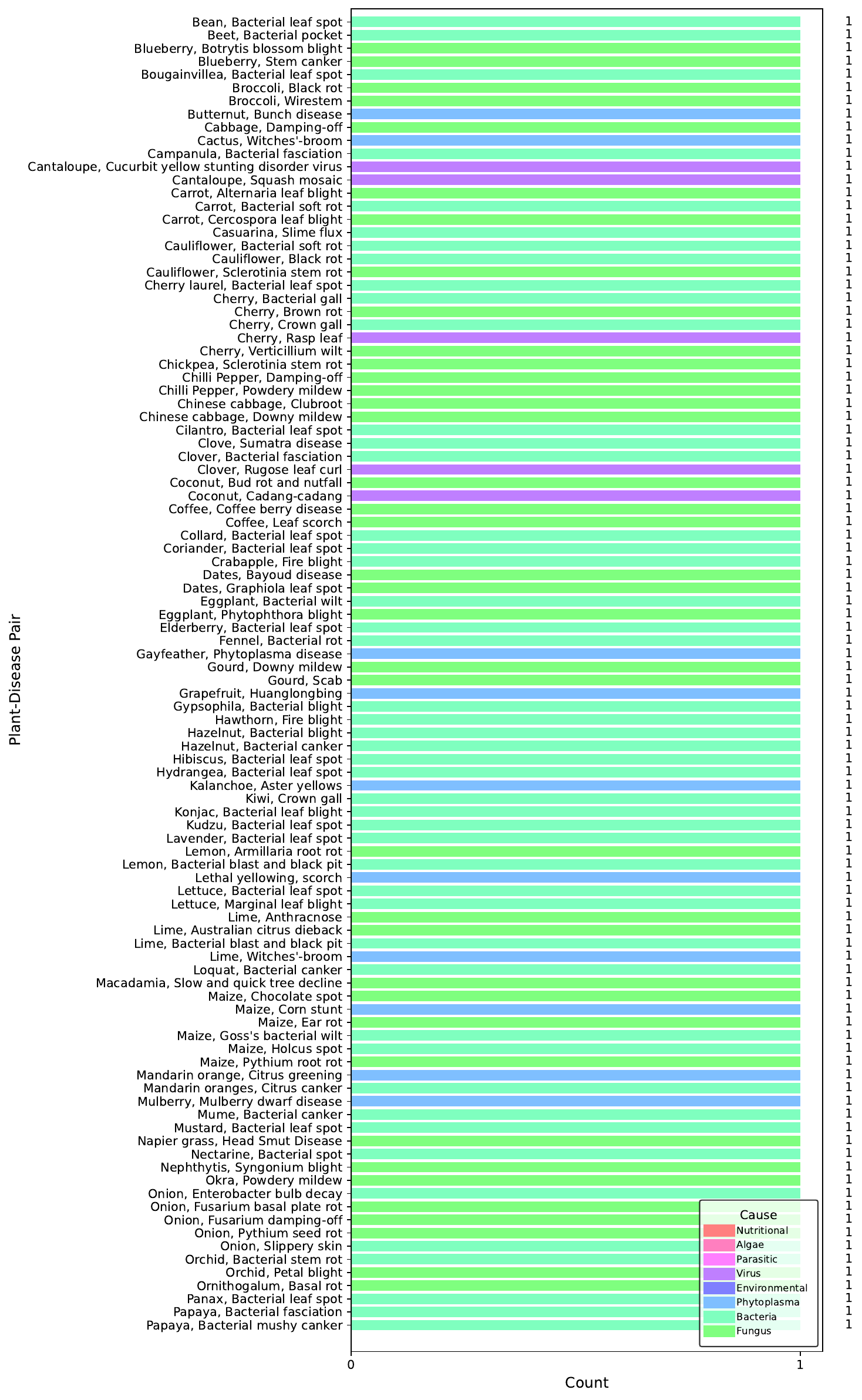}
\caption{\textbf{Disease Categories Counts (F).}}
\label{fig:diseasecountF}
\vspace{-11pt}
\end{figure*}

\begin{figure*}[ht]
\centering
\includegraphics[height=\textheight]{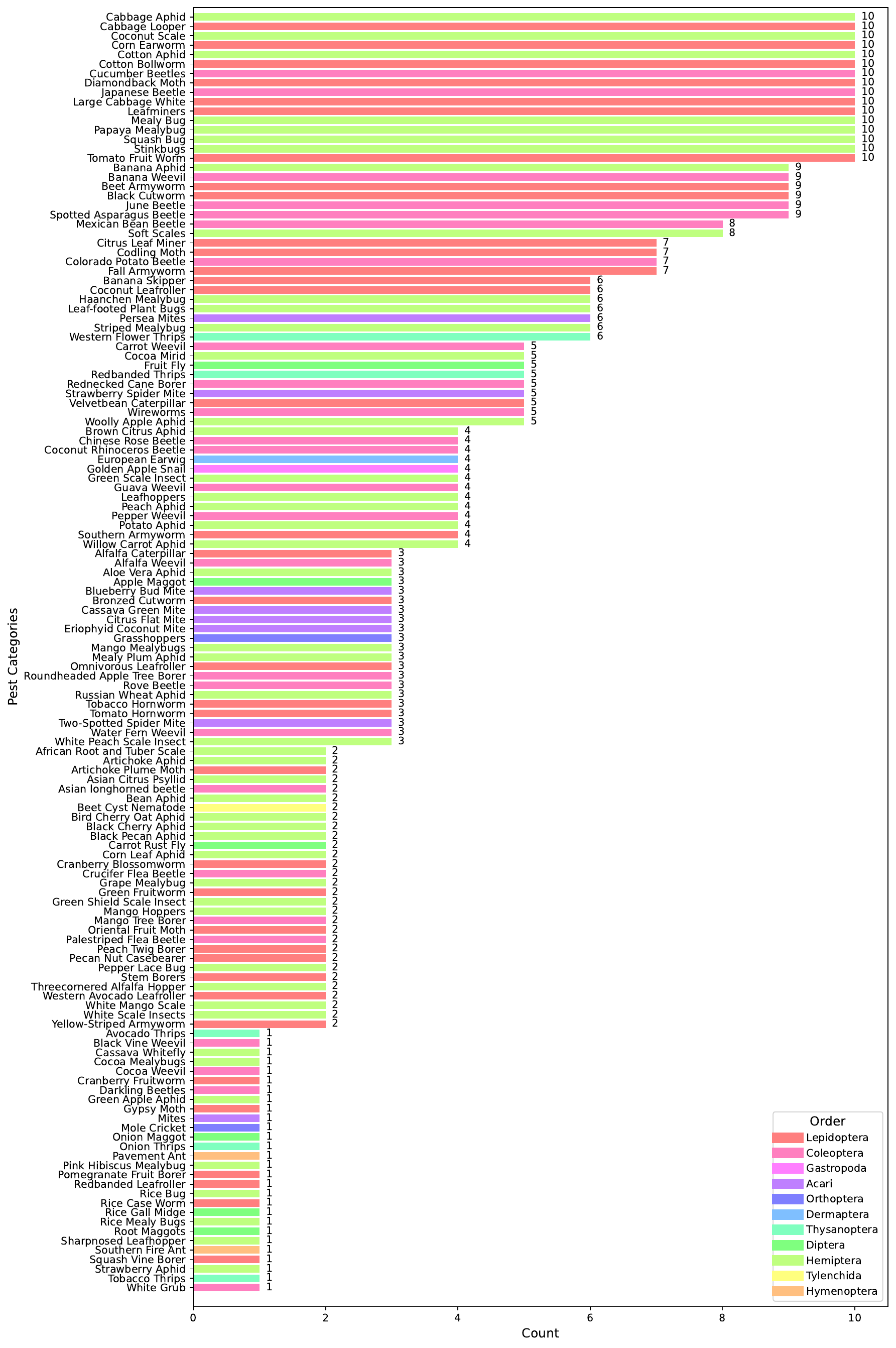}
\caption{\textbf{Pest Counts.} }
\label{fig:pestcount}
\vspace{-11pt}
\end{figure*}

\begin{figure*}[ht]
\centering
\includegraphics[height=\textheight]{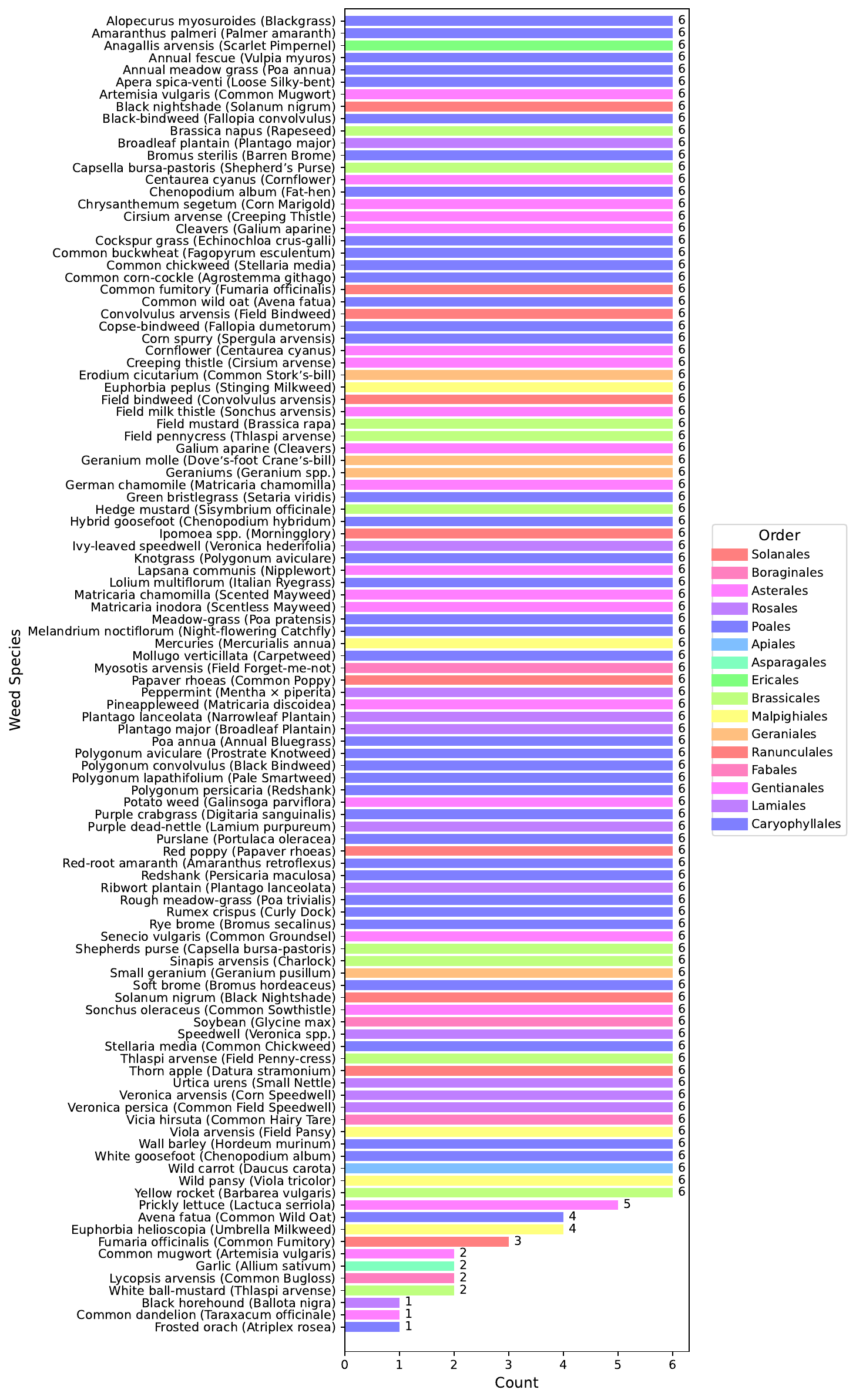}
\caption{\textbf{Weed Counts.} }
\label{fig:weedcount}
\vspace{-11pt}
\end{figure*}

This section provides detailed statistics of \datasetname.
In Figs. \ref{fig:diseasecountA}, \ref{fig:diseasecountB}, \ref{fig:diseasecountC}, \ref{fig:diseasecountD}, \ref{fig:diseasecountE}, and \ref{fig:diseasecountF}, we present the distribution of 682 plant-disease categories, categorized by the cause of the disease.
The distribution of 134 pest categories is shown in Fig. \ref{fig:pestcount}, and the distribution of 108 weed categories is shown in Fig. \ref{fig:weedcount}, both categorized by their order.

\section{Experimental Details}
\subsection{Hyperparameter Settings}
We follow the official hyperparameter settings. For closed-source models, we set the temperature hyperparameter to 0.0. For open-source models, we use the default temperature values provided by each implementation.
For GPT-4o and GPT-4o mini, we set the detail parameter for visual inputs to low.

\subsection{Model Details}
\label{sec:exdetails}
\noindent \textbf{GPT-4o~\cite{gpt4o} and GPT-4o mini~\cite{gpt4omini}.} GPT-4o is an extended model in the series of Generative Pre-training Transformers (GPTs). 
GPT-4o mini is a lightweight model designed for efficiency and speed. While it sacrifices a small degree of accuracy compared to GPT-4o, it remains an efficient option.
We use the model gpt-4o-2024-08-06 for GPT-4o and gpt-4o-mini-2024-07-18 for GPT-4o mini.

\noindent \textbf{Gemini 1.5-Pro and Gemini 1.5-Flash~\cite{gemini}.} Gemini has been simultaneously trained on a huge amount of multi-modality dataset among image, video, audio, and text data. 
Gemini 1.5-Flash is the lightweight version
of Gemini1.5-Pro. We use the model gemini-1.5-pro-001 for Gemini 1.5-Pro and gemini-1.5-flash-001 for Gemini 1.5-Flash.

\noindent \textbf{Qwen~\cite{qwen}.} Qwen is a vision-language model trained with a huge amount of image-text data and instruction tuning. Qwen has several specific modes, such as coding, additional audio modality, and mathematics. In this paper, we employ Qwen2-72B (QwenVLM-72B) and Qwen2-7B (QwenVLM-7B).

\noindent \textbf{LLaVA~\cite{liu2023llava}.} LLaVA takes advantage of trained large language models (LLMs) and instruction tuning with vision models. The LLaVA project has proved that the merged representation is greatly effective for visual reasoning and dialogue using multi-modal input. In the experiments, we use LLaVA v1.5 with 7 and 13 billion parameters (LLaVA-1.5-\{7B, 13B\}). We also use LLaVA-Next~\cite{liu2024llavanext} improved version of LLaVA.

\noindent \textbf{CogVLM~\cite{wang2023cogvlm}.} This vision-language model employs trained LLMs and visual encoders and additionally tunes feed-forward layers in order to combine the ability of image-text representations. CogVLM has confirmed question-answering and visual reasoning performance on representative vision and language datasets. We utilize the second version of CogVLM, which contains 19B parameters and has Llama-3 backbones.

\noindent \textbf{Emu~\cite{Emu}.} Emu potentially executes both image-to-text and text-to-image across diverse visual, linguistic, and multi-modal tasks. This foundation model enhances the vision and language performance from both recognition and generation learning across multiple modalities. The model size is 37B.

\section{Annotation Examples}
\begin{figure*}[t]
    \centering
    \includegraphics[width=0.9\textwidth]{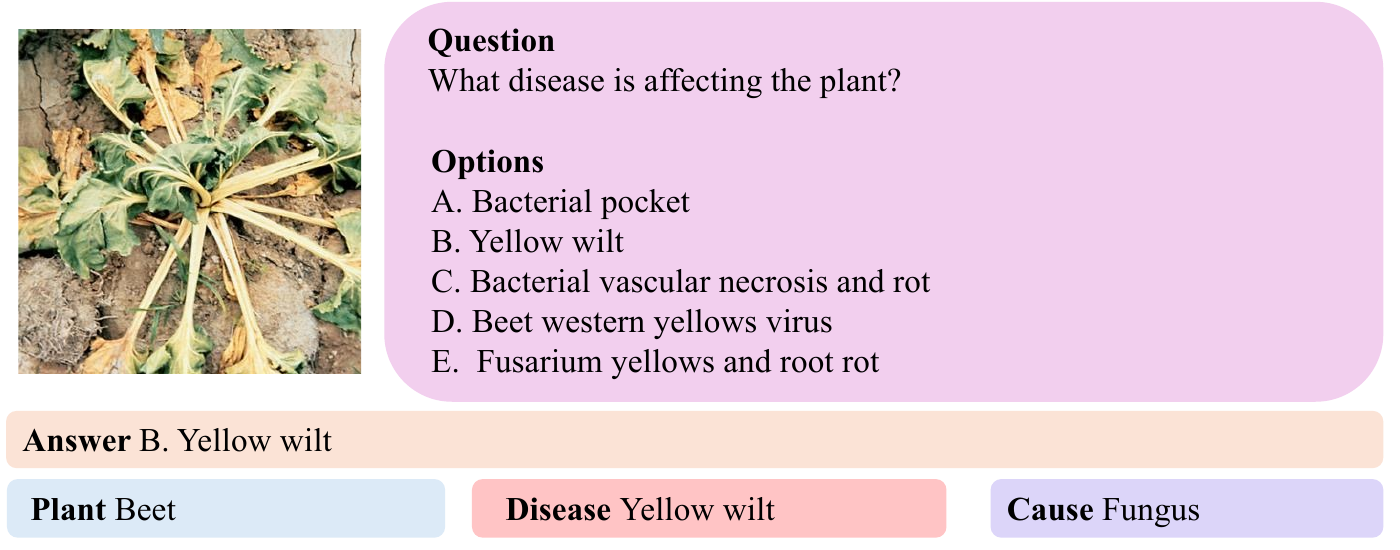}
  \caption{\textbf{Disease Identification Task (DID) Example (A).}}
\label{fig:didA}
\vspace{-11pt}  
\end{figure*}

\begin{figure*}[t]
    \centering
    \includegraphics[width=0.9\textwidth]{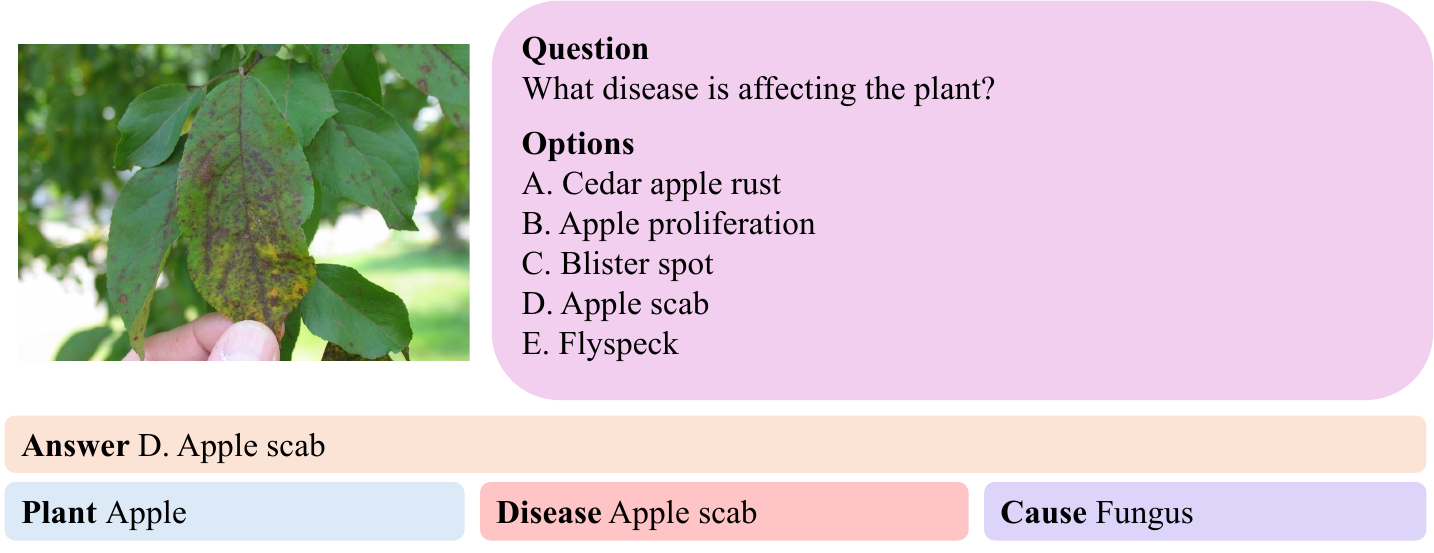}
\caption{\textbf{Disease Identification Task (DID) Example (B).}}
\label{fig:didB}
\vspace{-11pt}
\end{figure*}

\begin{figure*}[t]
    \centering
    \includegraphics[width=0.9\textwidth]{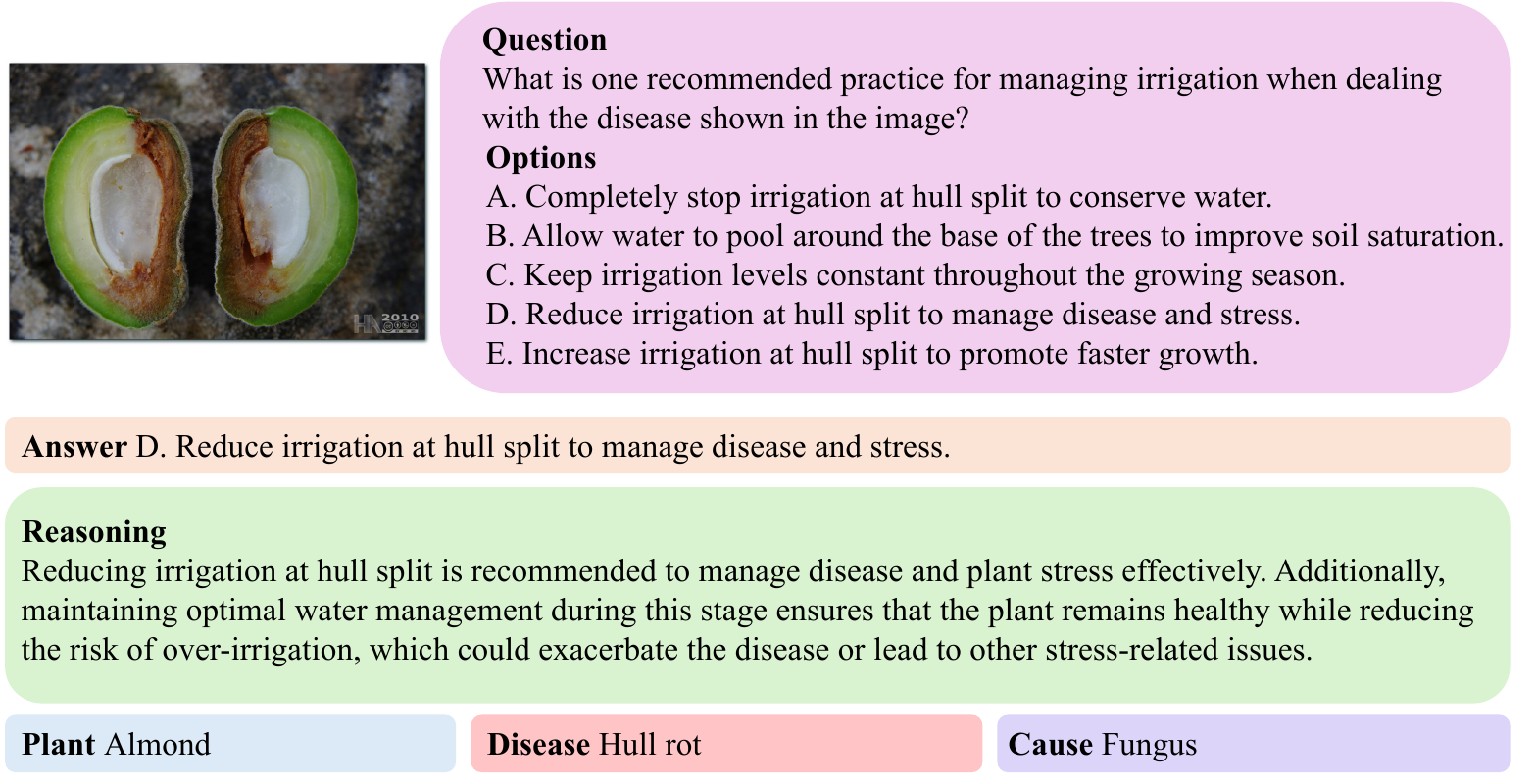}
  \caption{\textbf{Disease Management Task (DMN) Example (A).}}
\label{fig:dmnA}
\vspace{-11pt}  
\end{figure*}

\begin{figure*}[t]
    \centering
    \includegraphics[width=0.9\textwidth]{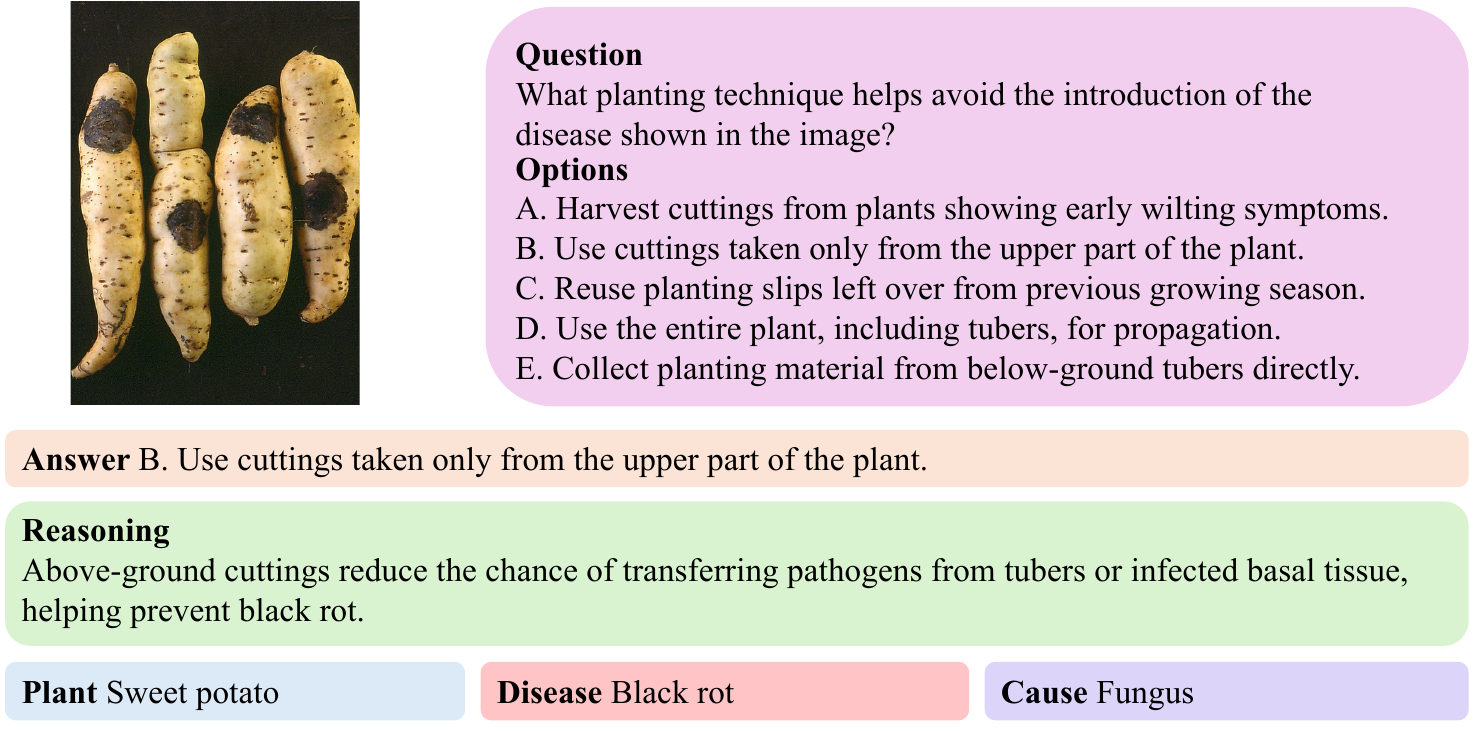}
\caption{\textbf{Disease Management Task (DMN) Example (B).}}
\label{fig:dmnB}
\vspace{-11pt}
\end{figure*}

\begin{figure*}[t]
    \centering
    \includegraphics[width=0.9\textwidth]{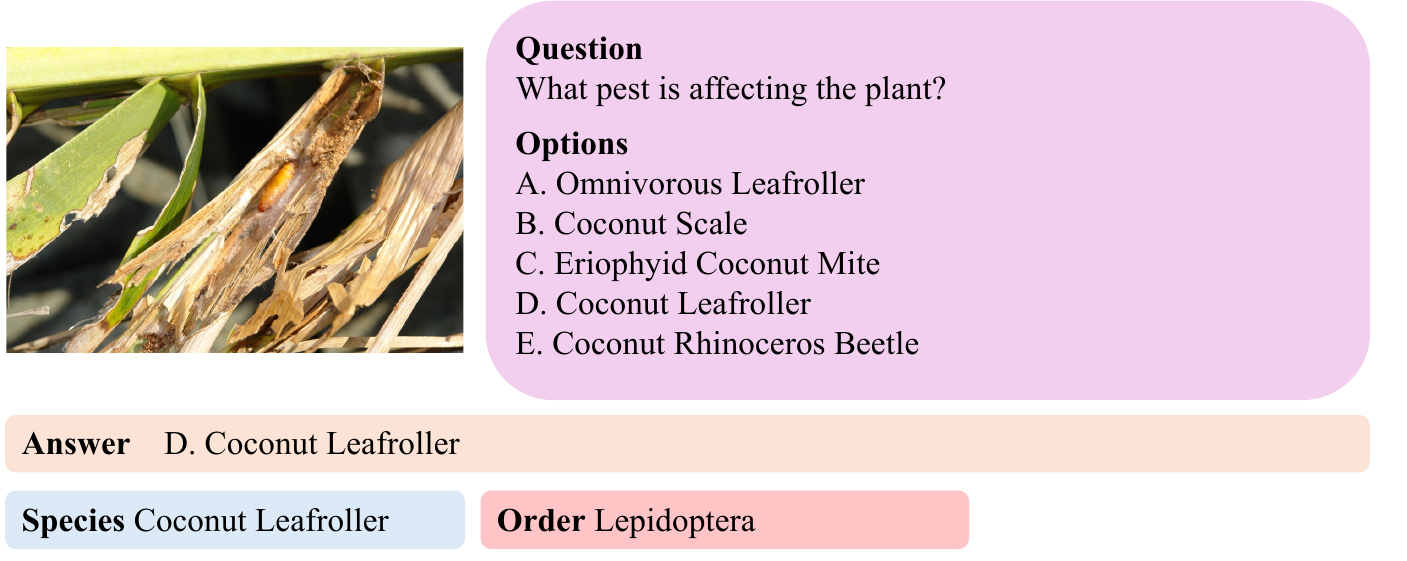}
  \caption{\textbf{Pest Identification Task (PID) Example (A).}}
\label{fig:pidA}
\vspace{-11pt}  
\end{figure*}

\begin{figure*}[t]
    \centering
    \includegraphics[width=0.9\textwidth]{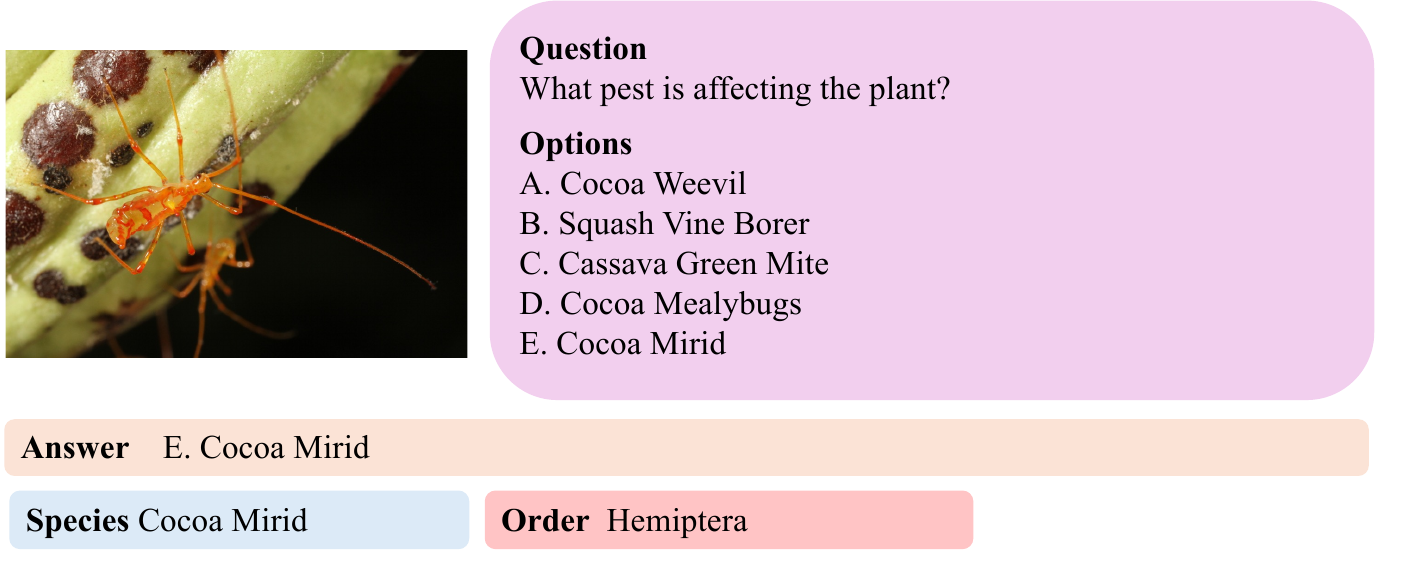}
\caption{\textbf{Pest Identification Task (PID) Example (B).}}
\label{fig:pidB}
\vspace{-11pt}
\end{figure*}

\begin{figure*}[t]
    \centering
    \includegraphics[width=0.9\textwidth]{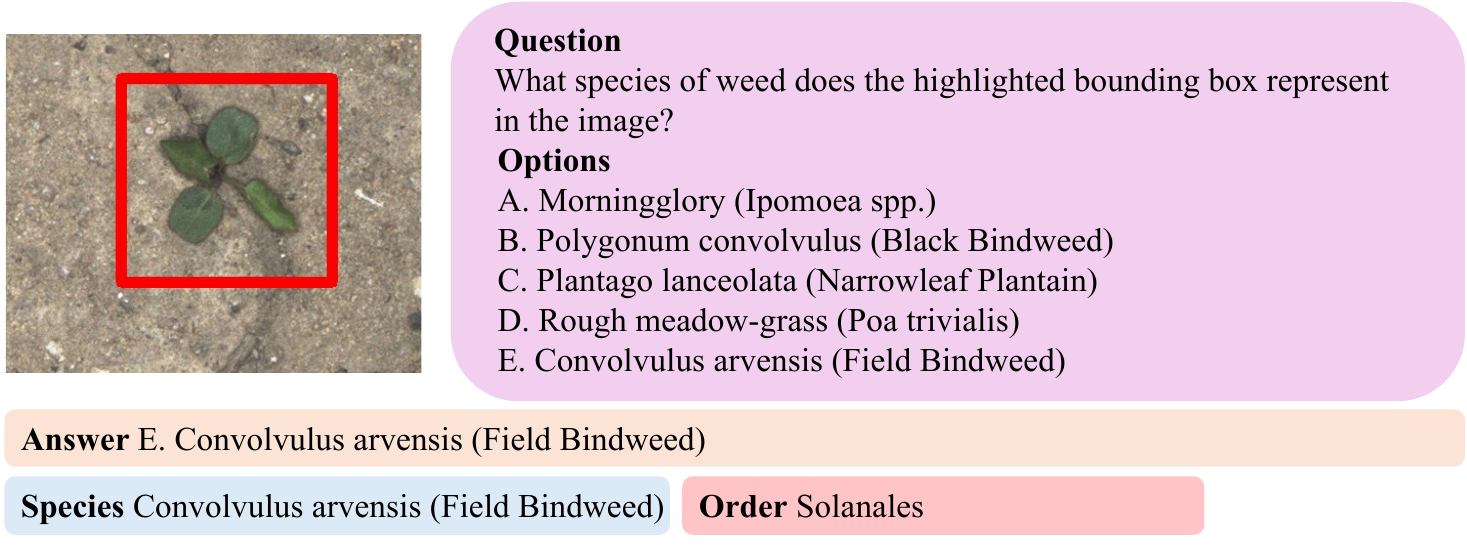}
  \caption{\textbf{Weed Identification Task (WID) Example (A).}}
\label{fig:widA}
\vspace{-11pt}  
\end{figure*}

\begin{figure*}[t]
    \centering
    \includegraphics[width=0.9\textwidth]{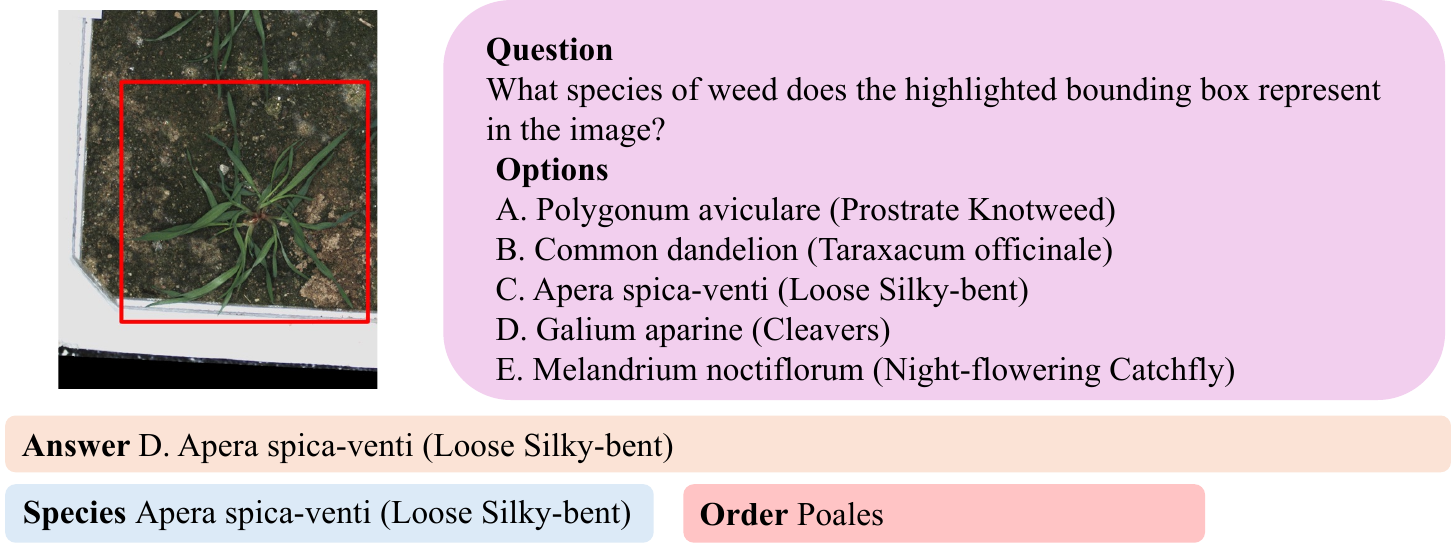}
\caption{\textbf{Weed Identification Task (WID) Example (B).}}
\label{fig:widB}
\vspace{-11pt}
\end{figure*}

\begin{figure*}[t]
    \centering
    \includegraphics[width=0.9\textwidth]{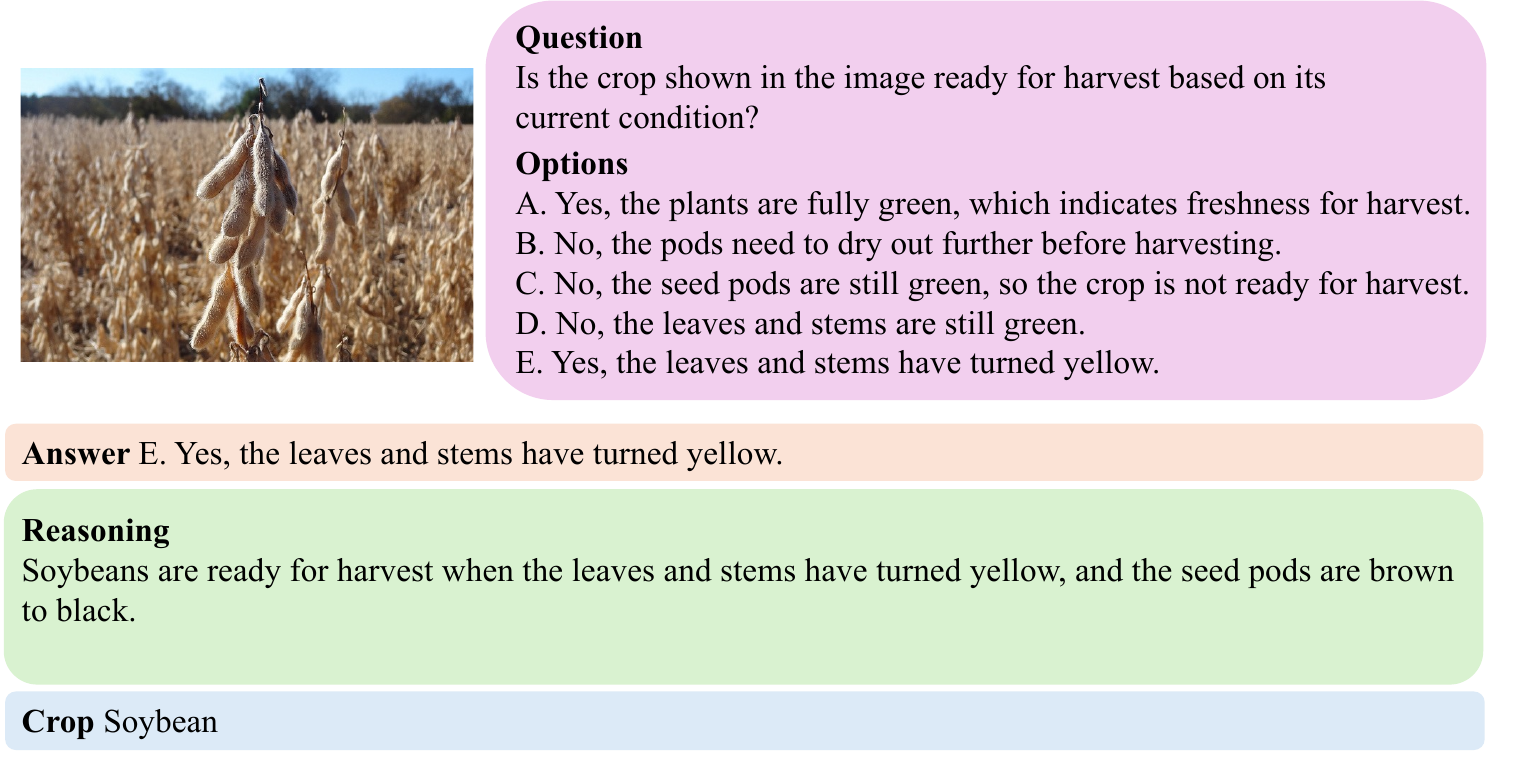}
  \caption{\textbf{Crop Management Task (CMN) Example (A).}}
\label{fig:cmnA}
\vspace{-11pt}  
\end{figure*}

\begin{figure*}[t]
    \centering
    \includegraphics[width=0.9\textwidth]{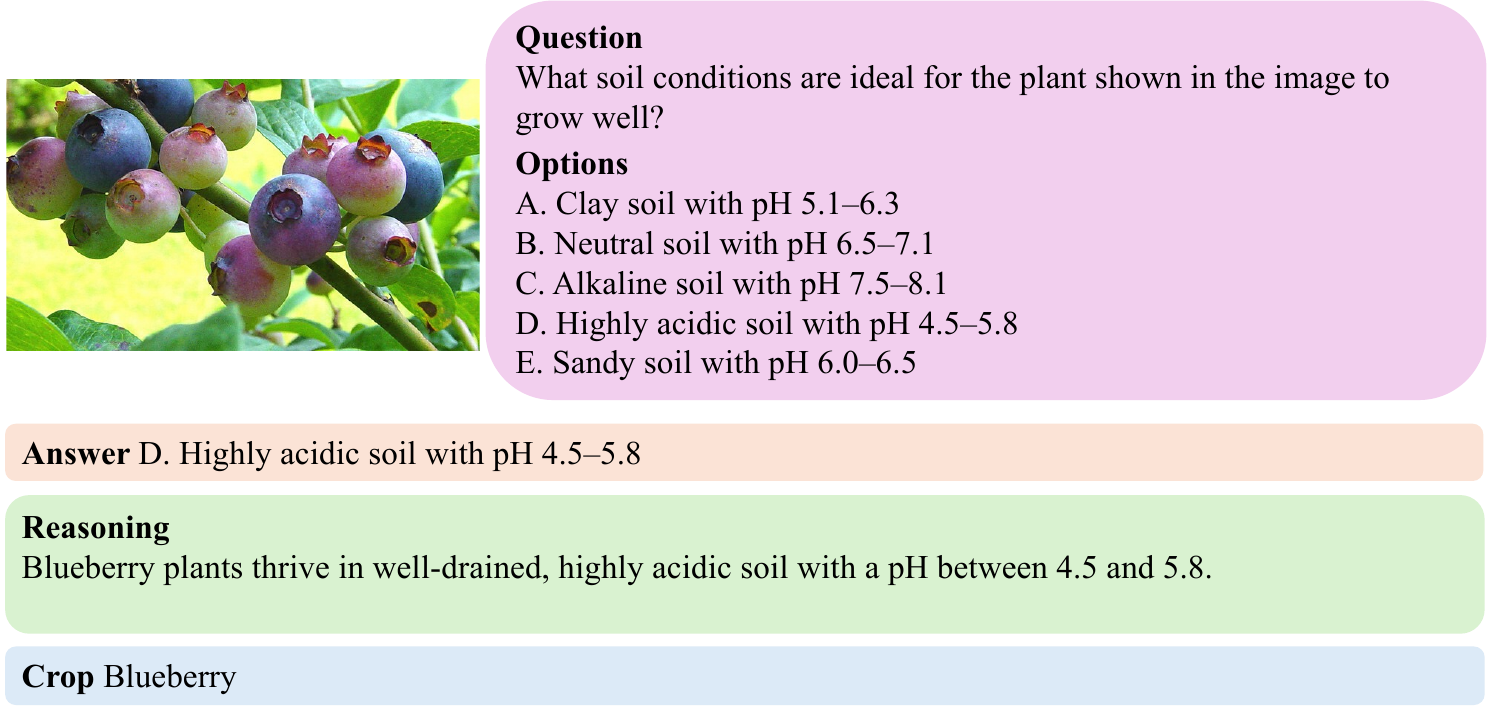}
\caption{\textbf{Crop Management Task (CMN) Example (B).}}
\label{fig:cmnB}
\vspace{-11pt}
\end{figure*}

\begin{figure*}[t]
    \centering
    \includegraphics[width=0.9\textwidth]{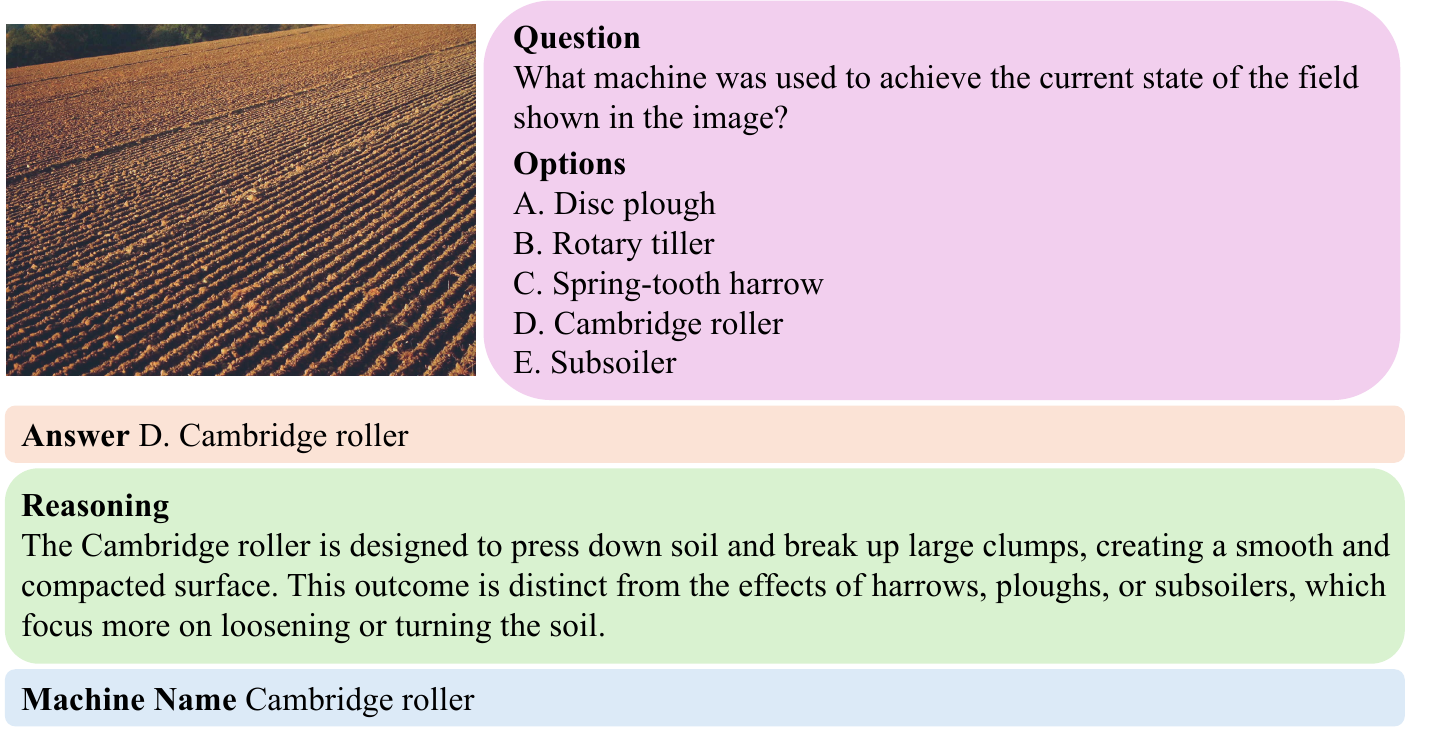}
  \caption{\textbf{Machine Usage QA Task (MQA) Example (A).}}
\label{fig:mqaA}
\vspace{-11pt}  
\end{figure*}

\begin{figure*}[t]
    \centering
    \includegraphics[width=0.9\textwidth]{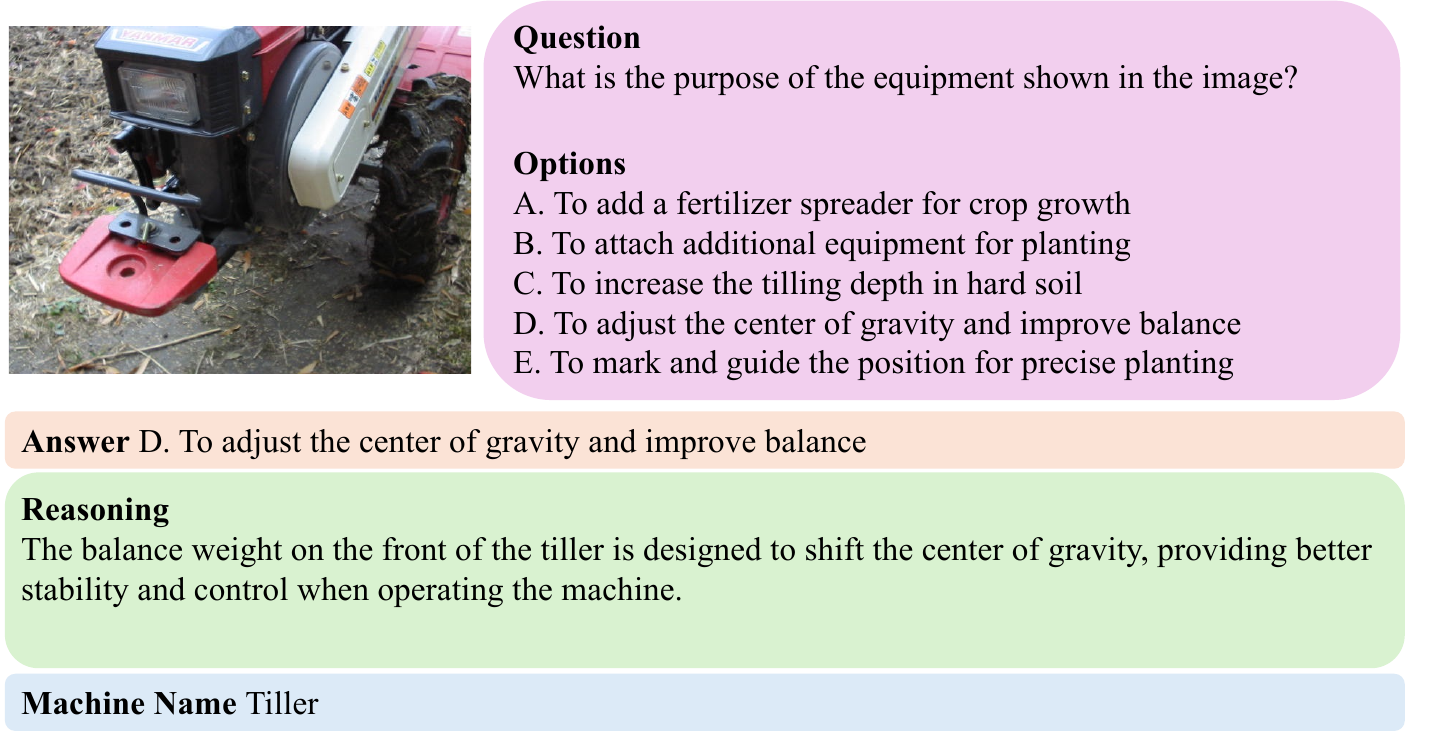}
\caption{\textbf{Machine Usage QA Task (MQA) Example (B).}}
\label{fig:mqaB}
\vspace{-11pt}
\end{figure*}

\begin{figure*}[t]
    \centering
    \includegraphics[width=0.9\textwidth]{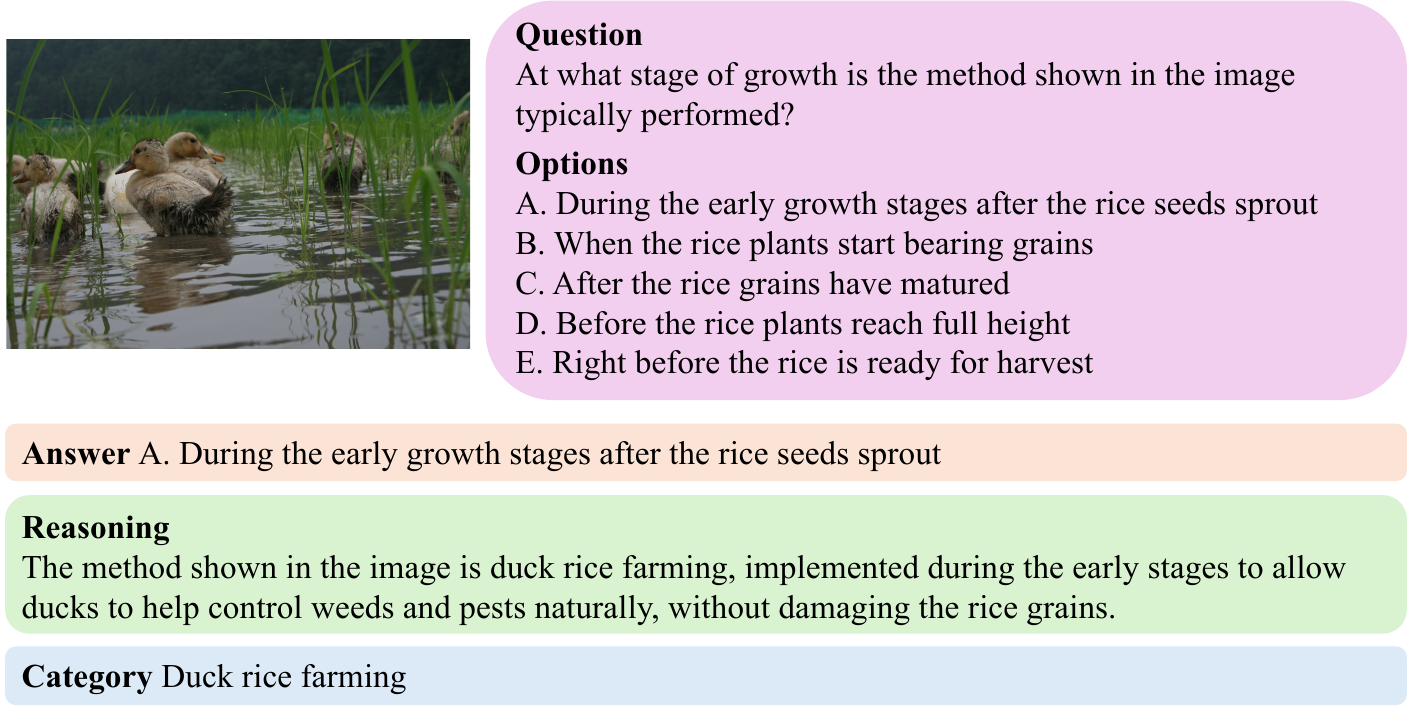}
  \caption{\textbf{Traditional Methods (TM) Example (A).}}
\label{fig:tmA}
\vspace{-11pt}  
\end{figure*}

\begin{figure*}[t]
    \centering
    \includegraphics[width=0.9\textwidth]{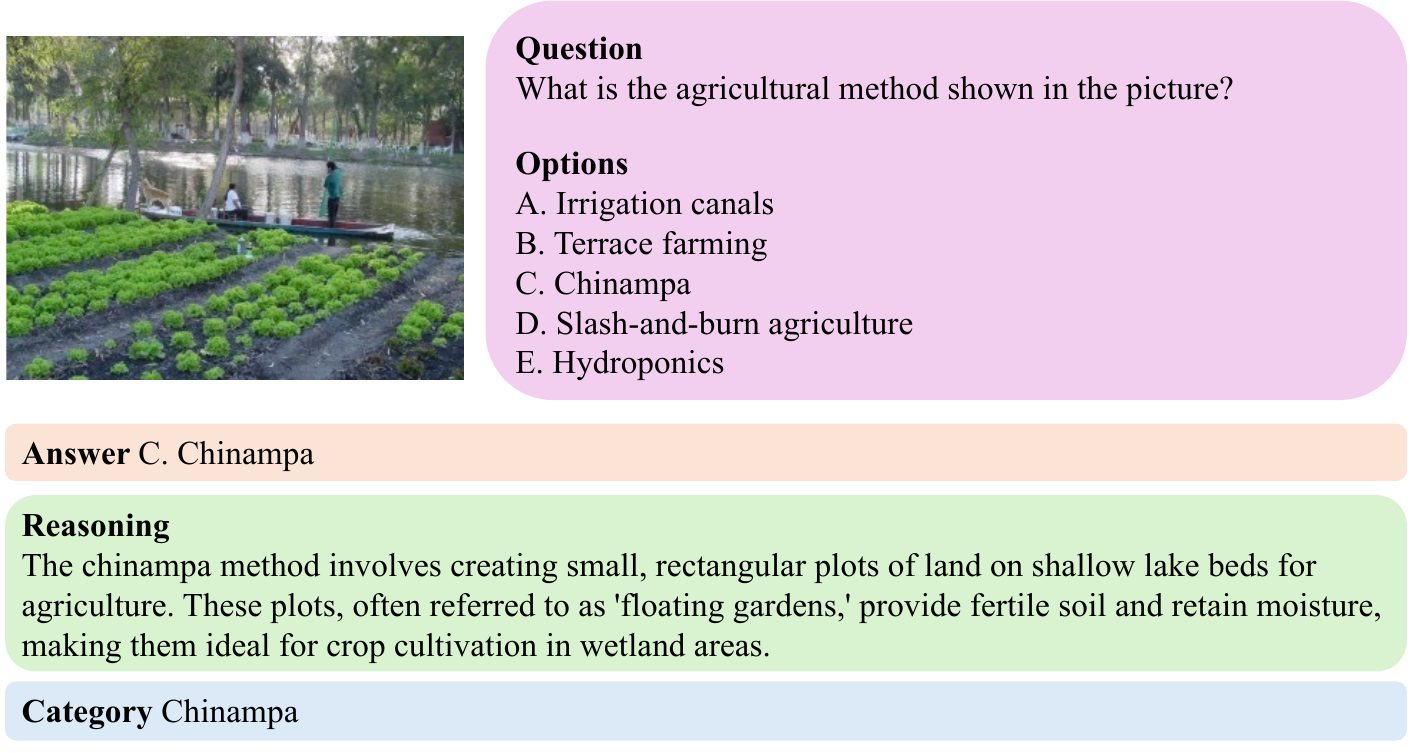}
\caption{\textbf{Traditional Methods QA (TM) Example (B).}}
\label{fig:tmB}
\vspace{-11pt}
\end{figure*}

In this section, we show example annotations for the seven tasks as follows:
\begin{itemize}
    \item Disease Identification~(DID): Fig. \ref{fig:didA}, Fig. \ref{fig:didB}
    \item Disease Management~(DMN): Fig. \ref{fig:dmnA}, Fig. \ref{fig:dmnB}
    \item Pest Identification~(PID): Fig. \ref{fig:pidA}, Fig. \ref{fig:pidB}
    \item Weed Identification~(WID): Fig. \ref{fig:widA}, Fig. \ref{fig:widB}
    \item Crop Management~(CMN): Fig. \ref{fig:cmnA}, Fig. \ref{fig:cmnB}
    \item Machine Usage~(MQA): Fig. \ref{fig:mqaA}, Fig. \ref{fig:mqaB}
    \item Traditional Methods~(TM): Fig. \ref{fig:tmA}, Fig. \ref{fig:tmB}
\end{itemize}

As shown in the figures, we provide reasoning annotations for DMN, CMN, MQA, and TM tasks, ensuring explainability. We also list the category for TM in Table.~\ref{tab:traditional_methods} and for MQA in Table.~\ref{tab:machinery_categories}.
\section{Detailed Analysis}
\subsection{Context Reliance}
\begin{figure*}[t]
    \centering
    \includegraphics[width=0.9\textwidth]{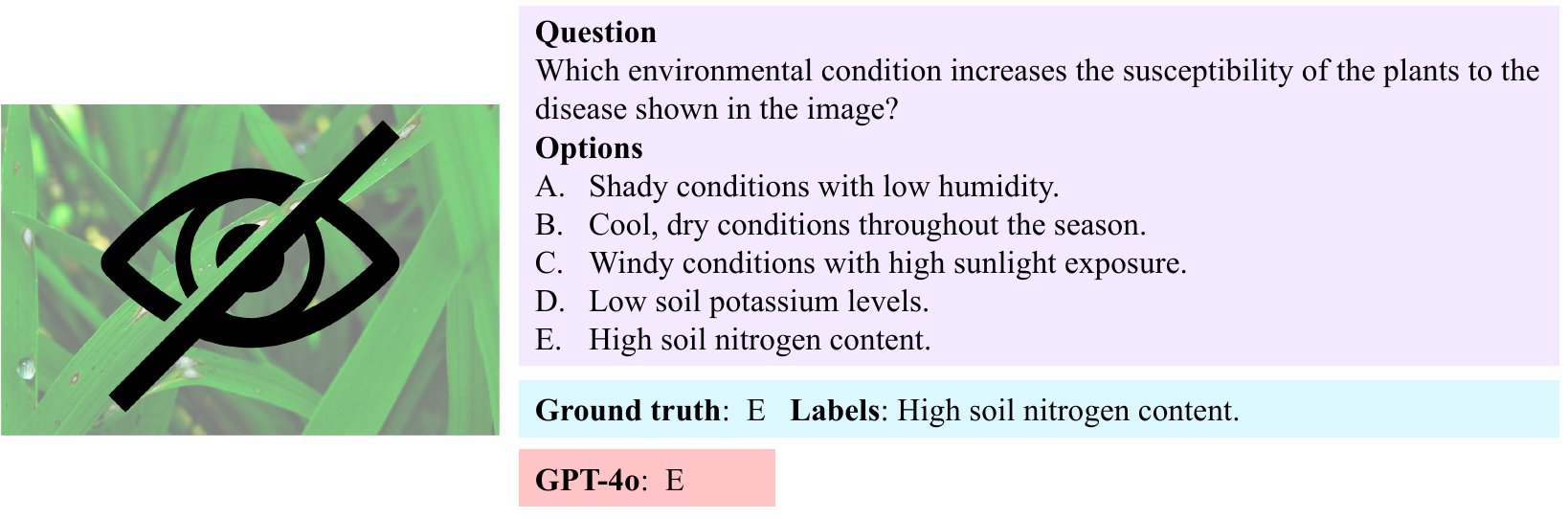}
  \caption{\textbf{Example case of GPT-4o answering without an input image (DMN).}
Even when the model cannot determine the crop or disease based on the text, it guesses the answer.}
\label{fig:contA}
\vspace{-11pt}  
\end{figure*}

\begin{figure*}[t]
    \centering
    \includegraphics[width=0.9\textwidth]{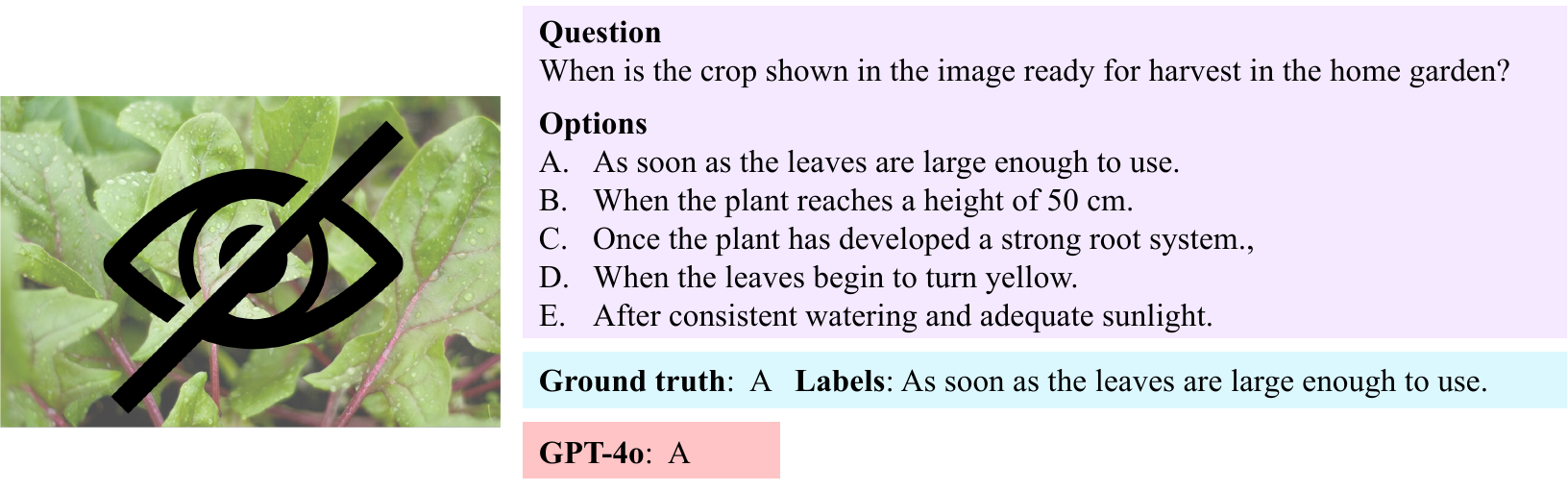}
\caption{\textbf{Example case of GPT-4o answering without an input image (CMN).}
Even when the model cannot determine the crop based on the text, it guesses the answer.}
\label{fig:contB}
\vspace{-11pt}
\end{figure*}

We show example cases where GPT-4o can answer questions without input images by guessing the most likely option~ (Fig. \ref{fig:contA} and Fig. \ref{fig:contB}). Even though we confirmed that our prepared questions do not include crop names or appearance traits, and other options could be correct, the models tend to make conclusions based on estimation.

\subsection{CoT Examples}
\begin{figure*}[t]
    \centering
    \includegraphics[width=0.9\textwidth]{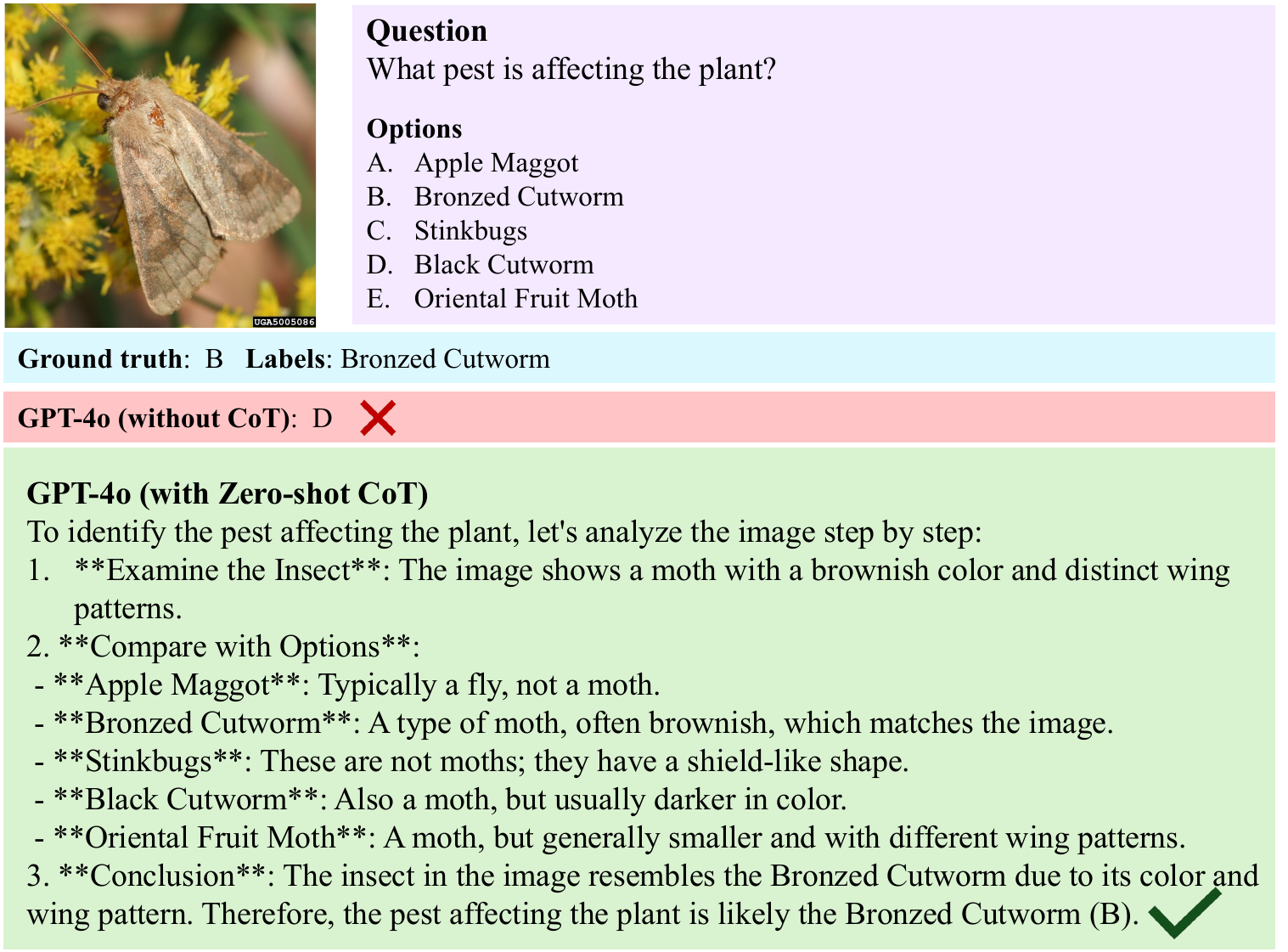}
\caption{\textbf{Example case of zero-shot CoT success.} The model identifies the pest step by step by observing the image's appearance, checks all the options, and concludes with the correct answer.}
\label{fig:cot_zero}
\vspace{-11pt}
\end{figure*}

\begin{figure*}[t]
    \centering
    \includegraphics[width=0.9\textwidth]{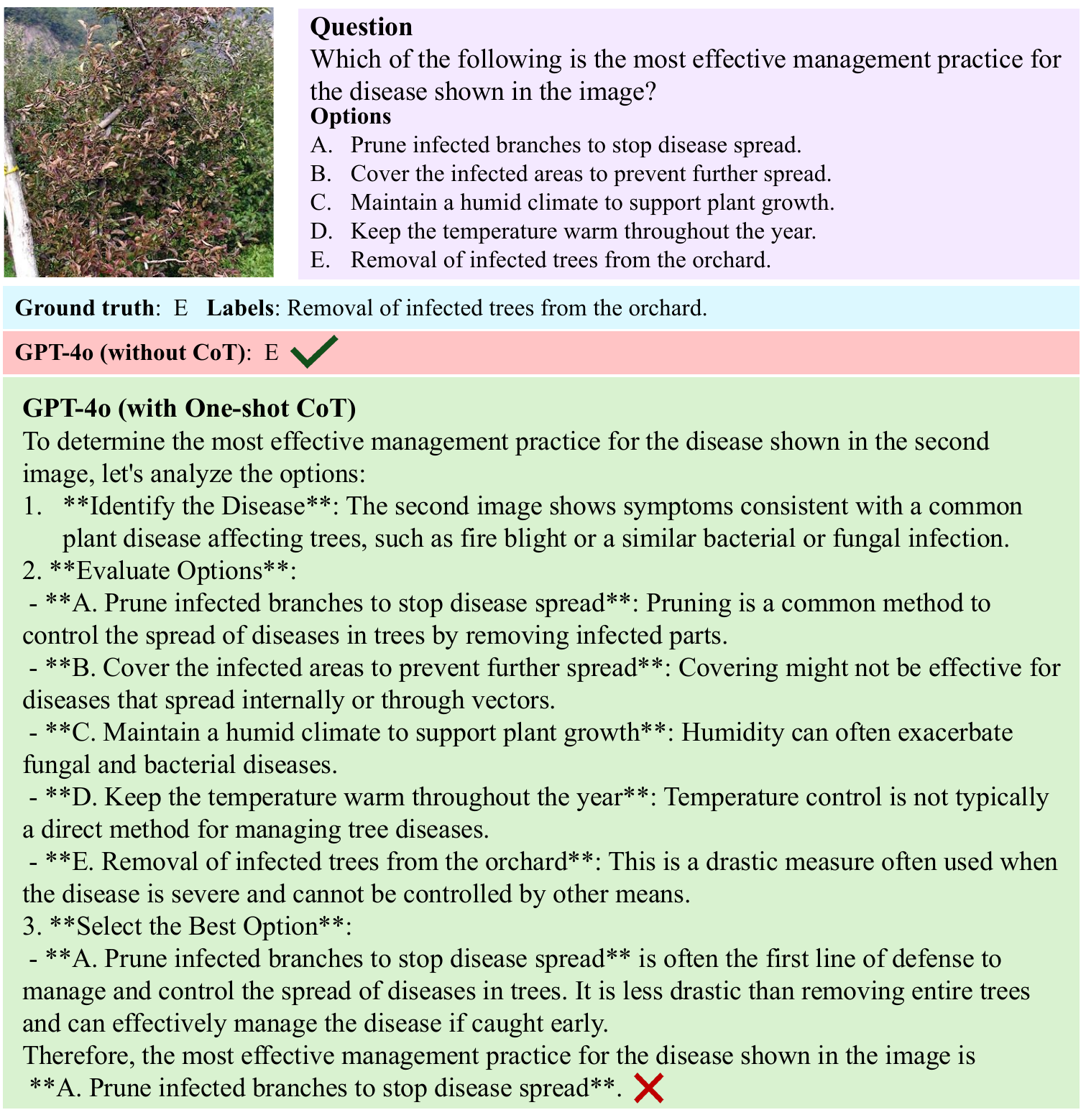}
\caption{\textbf{Example of one-shot CoT failure.} The model focuses on determining the plant disease species but fails to observe the severity of the disease, leading to an incorrect conclusion.}
\label{fig:cot_one}
\vspace{-11pt}
\end{figure*}

Here, we show output examples of chain-of-thought (CoT) reasoning by GPT-4o.
In Fig. \ref{fig:cot_zero}, the model identifies the pest step by step by observing the image’s appearance,
checks all the options, and concludes with the correct answer.
In Fig. \ref{fig:cot_one}, the model focuses on determining the plant disease species but fails to observe the severity of the disease, leading to an incorrect conclusion.

\subsection{Error Examples}
\begin{figure*}[t]
    \centering
    \includegraphics[width=0.9\textwidth]{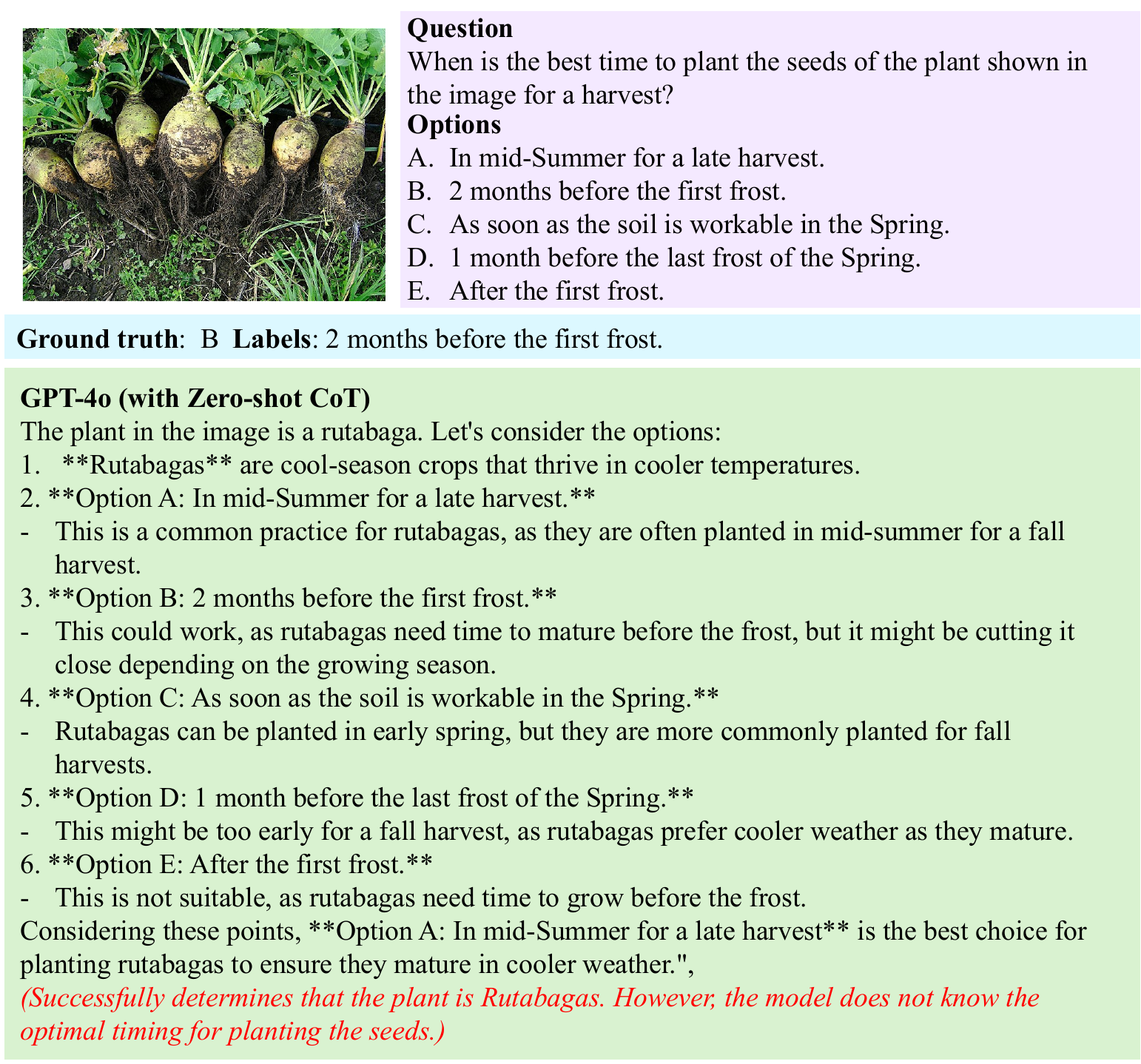}
\caption{\textbf{Example of Lack of knowledge Error (CMN).} The model successfully determines that the plant is Rutabagas. However, the model does not know the optimal timing for planting the seeds.}
\label{fig:inc_lack}
\vspace{-11pt}
\end{figure*}

\begin{figure*}[t]
    \centering
    \includegraphics[width=0.9\textwidth]{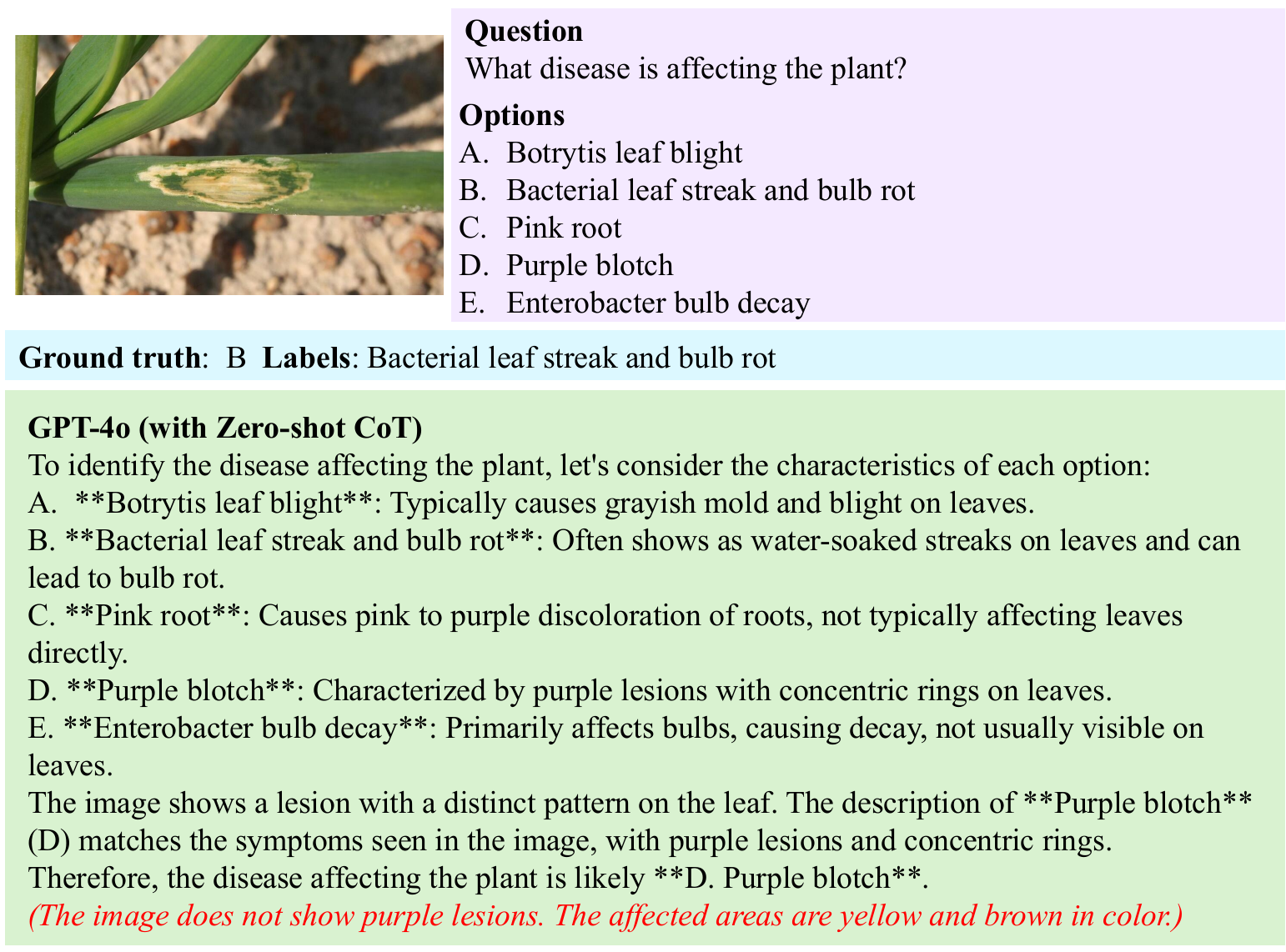}
\caption{\textbf{Example of Perceptual Error (DID).} The image shows the bacterial leaf streak and bulb rot, and the correct answer is B. The image does not show purple lesions. The affected areas are yellow and brown.}
\label{fig:inc_visu}
\vspace{-11pt}
\end{figure*}

\begin{figure*}[t]
    \centering
    \includegraphics[width=0.9\textwidth]{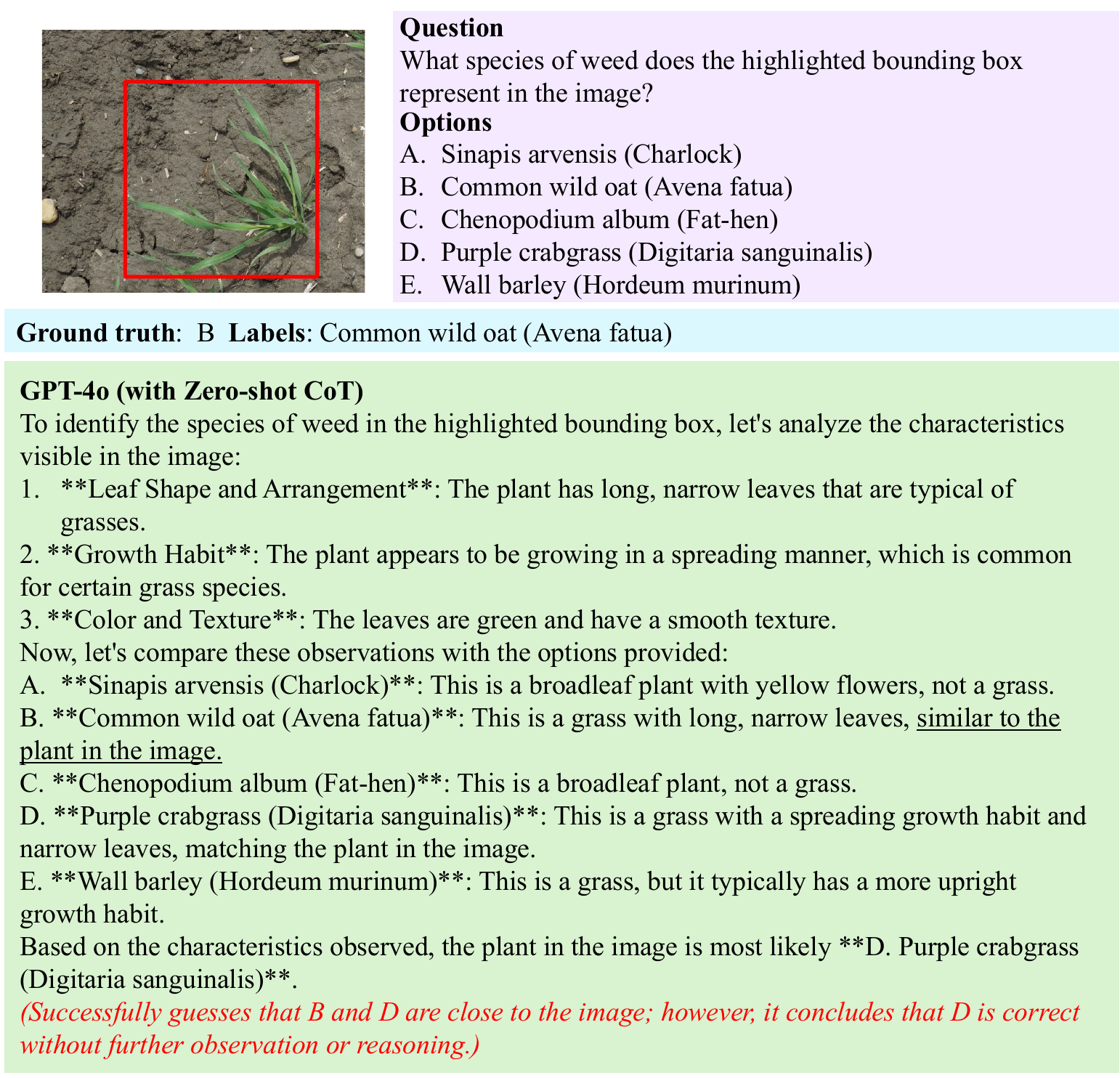}
\caption{\textbf{Example of Perceptual Error (DID).} The image shows the Common wild oat (Avena fatua), and the correct answer is B. The model successfully guesses that B and D are close to the image; however, it concludes that D is correct without further observation or reasoning.}
\label{fig:inc_short}
\vspace{-11pt}
\end{figure*}

\begin{figure*}[t]
    \centering
    \includegraphics[width=0.9\textwidth]{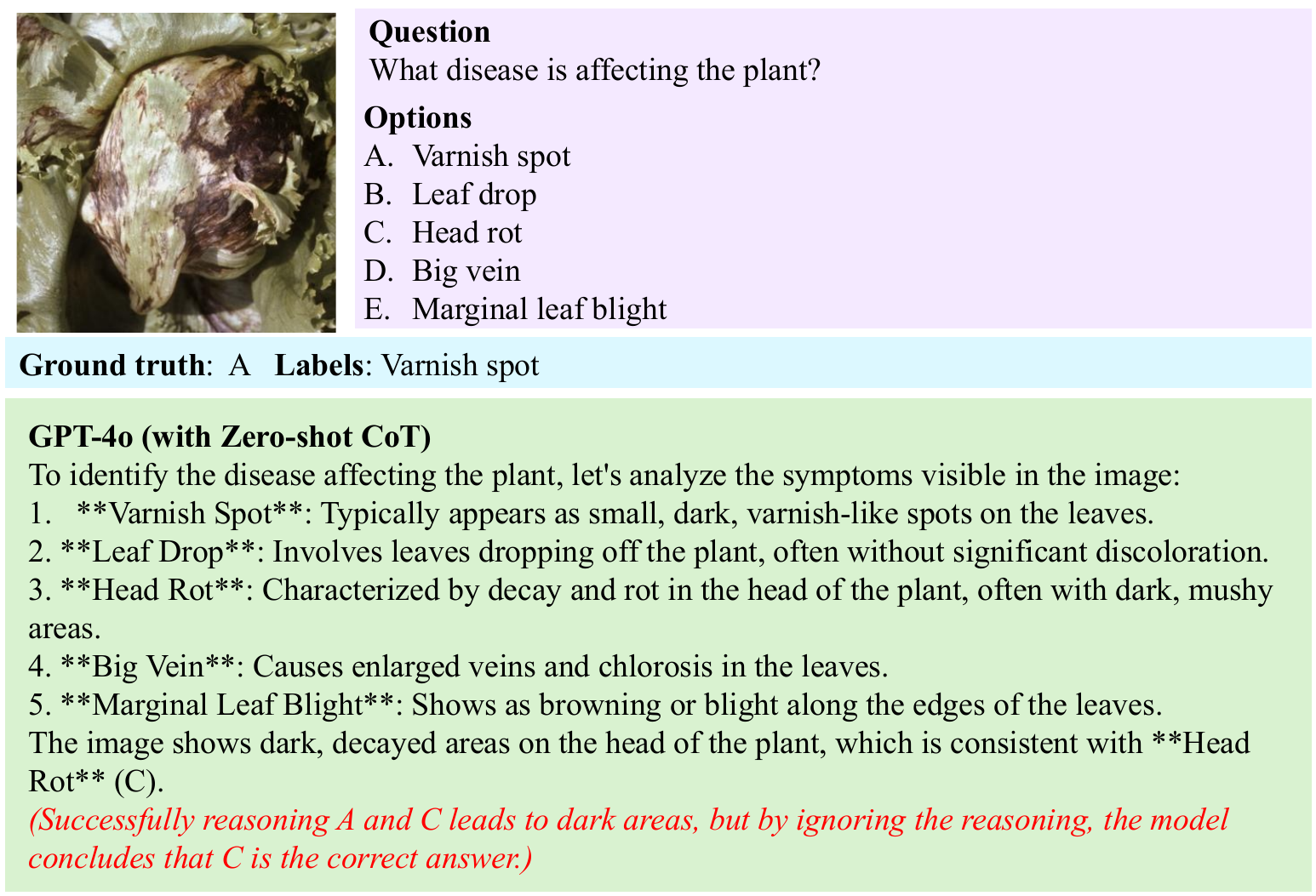}
\caption{\textbf{Example of Reasoning Error (DID).} Successfully reasoning A and C leads to dark areas, but by ignoring the reasoning, the model concludes that C is the correct answer.)}
\label{fig:inc_reasoning}
\vspace{-11pt}
\end{figure*}

\begin{figure*}[t]
    \centering
    \includegraphics[width=0.9\textwidth]{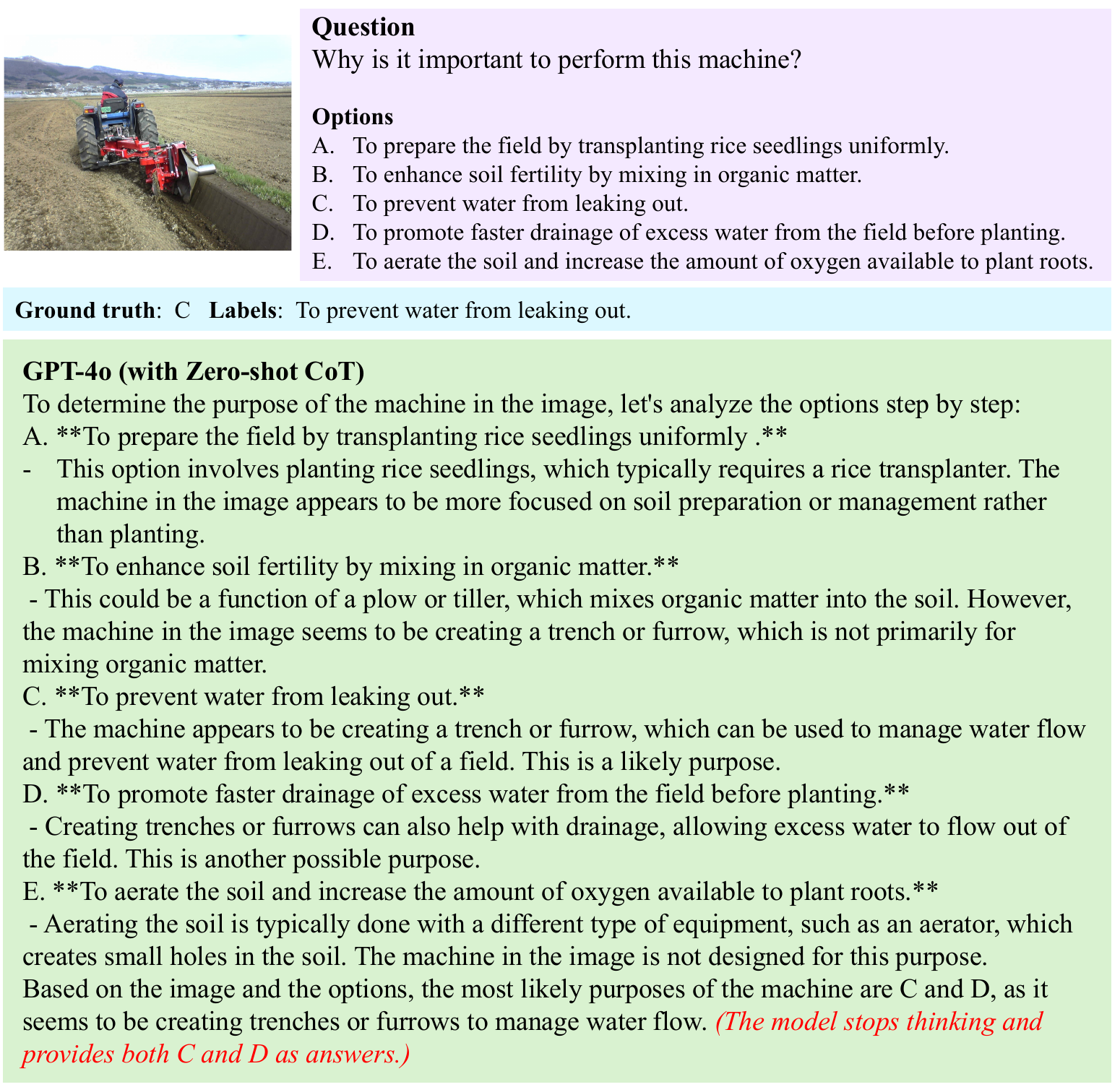}
\caption{\textbf{Example of Double Answer Error (MQA).} The image shows the border coating machine with the correct answer: C. However, the model stops thinking and provides both C and D as answers, even though it should continue reasoning to determine a single correct answer.}
\label{fig:inc_double}
\vspace{-11pt}
\end{figure*}

\begin{figure*}[t]
    \centering
    \includegraphics[width=0.9\textwidth]{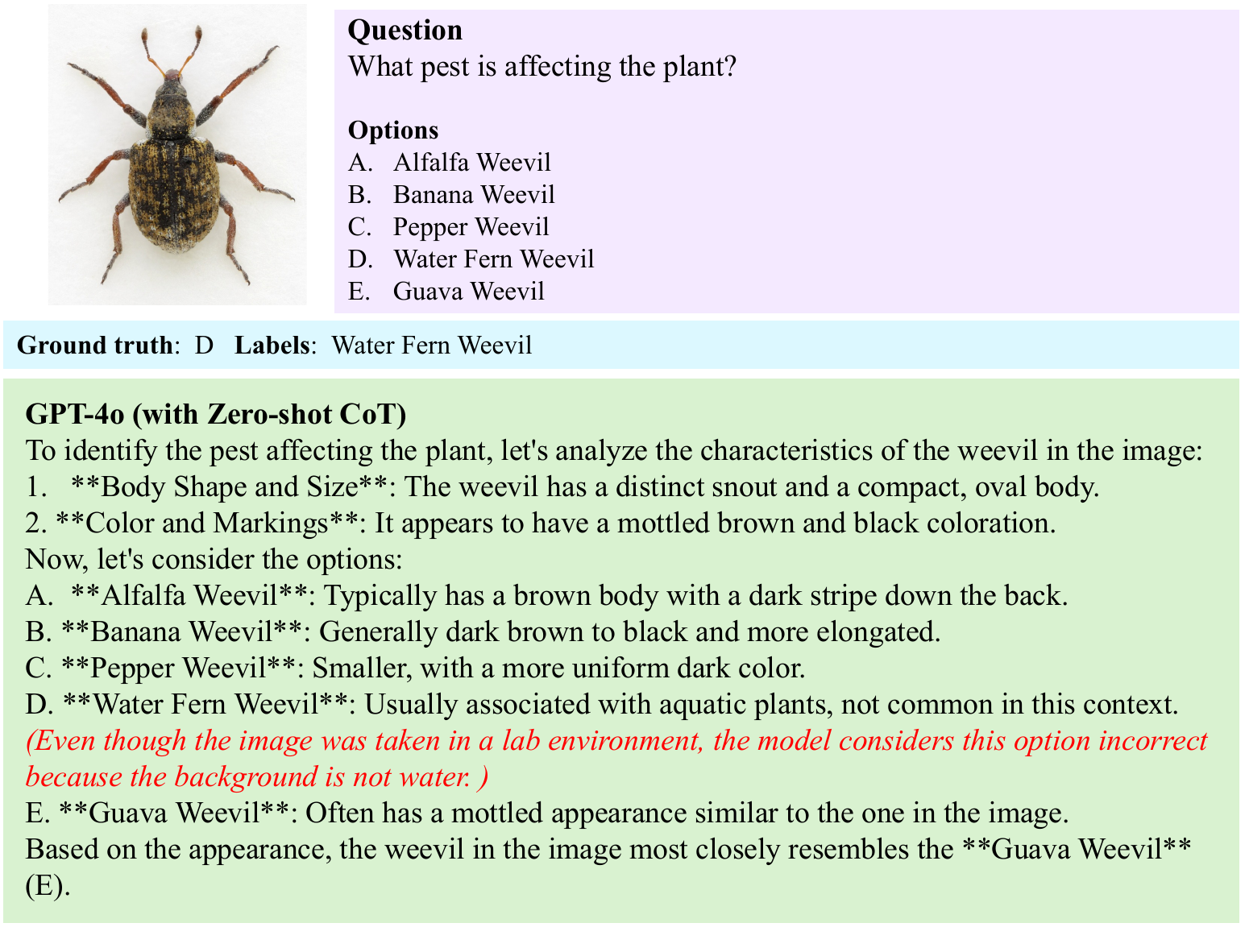}
\caption{\textbf{Example of Interpretation Misunderstanding (PID).} The image shows the Water Fern Weevil, and the correct answer is D. However, even though the image was taken in a lab environment, the model considers this option incorrect because the background is not water.}
\label{fig:inc_inter}
\vspace{-11pt}
\end{figure*}

\begin{figure*}[t]
    \centering
    \includegraphics[width=0.9\textwidth]{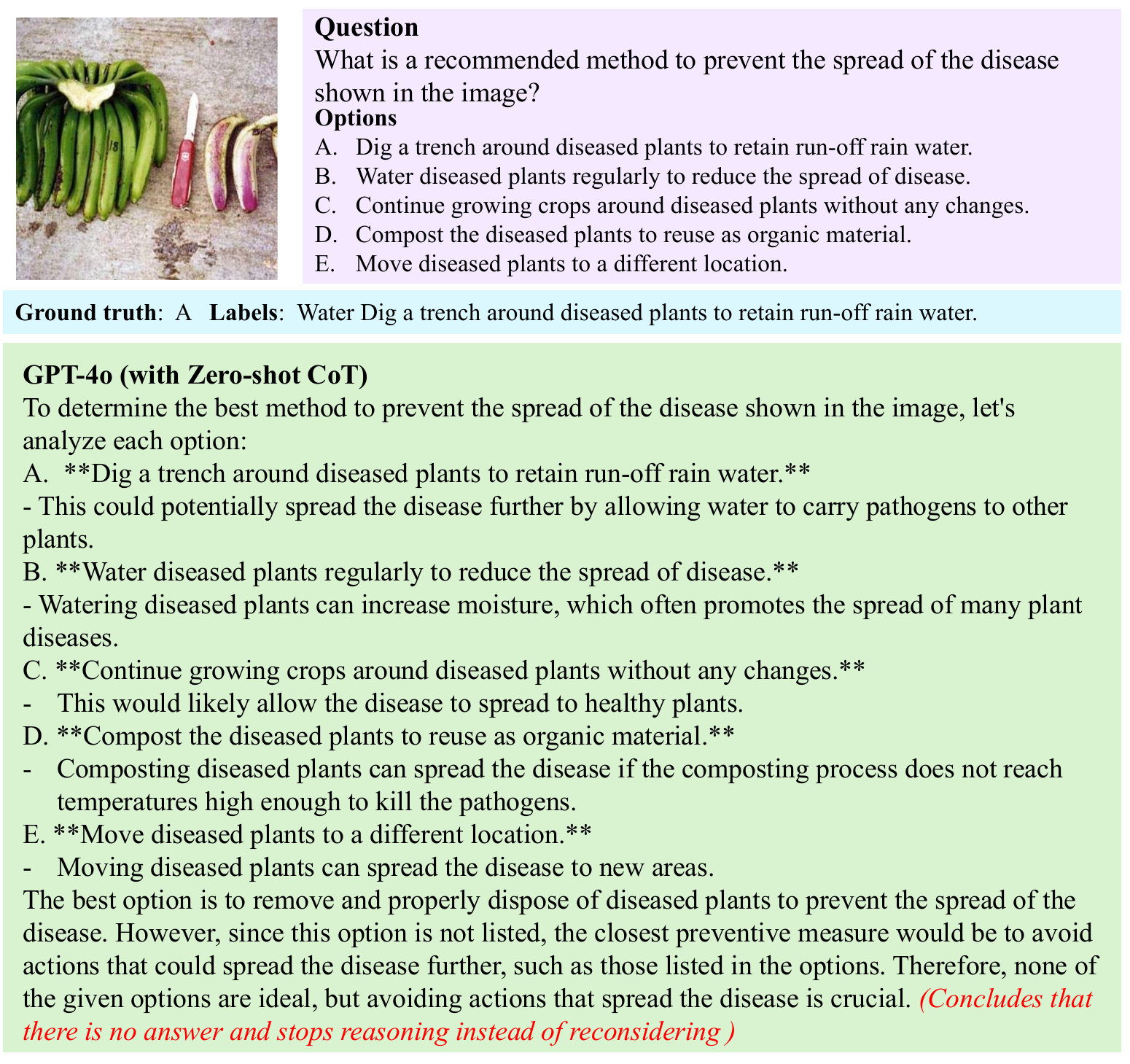}
\caption{\textbf{Example of Reject to Answer (DMN).} The image shows the blood disease of the banana, and the correct answer is A. However, the model concludes that there is no answer and stops reasoning instead of reconsidering.}
\label{fig:inc_reject}
\vspace{-11pt}
\end{figure*}

Here, we present additional error examples as follows:
\begin{itemize}
    \item Lack of Knowledge Error: Fig. \ref{fig:inc_lack}
    \item Perceptual Error: Fig. \ref{fig:inc_visu}
    \item Shortcut Error: Fig. \ref{fig:inc_short}
    \item Reasoning Error: Fig. \ref{fig:inc_reasoning}
    \item Double Answer Error: Fig. \ref{fig:inc_double}
    \item Interpretation Misunderstanding: Fig. \ref{fig:inc_inter}
    \item Reject to Answer: Fig. \ref{fig:inc_reject}
\end{itemize}

\subsection{Free format}
We evaluate our dataset under a free-form answer setting, where the model is not constrained to choose from multiple choices.
For the identification tasks (DID, PID, WID), we use the following prompt:
\begin{quote}
\texttt{\{question\} Respond with the correct answer only, using a single noun or short phrase. Do not include full sentences.}
\end{quote}
We show the result in Table~\ref{tab:f1}. The responses are evaluated using the F1 score and substring match accuracy, which reflect surface-level correctness as well as partial matches.

For the other tasks (DMN, CMN, MQA, TM), the prompt used is:
\begin{quote}
\texttt{\{question\} Respond with the correct answer only.}
\end{quote}
We show the result in Table~\ref{tab:bert}. The answers are evaluated with two semantic similarity metrics: BERTScore (F1) and Sentence-BERT cosine similarity.
\begin{table}[t]
    \centering
\begin{tabular}{lccc}
\toprule
Metric & DID & PID & WID \\
\midrule
F1 & 27.26 & 27.09 & 1.84 \\
Substring & 24.57 & 30.33 & 9.03 \\
\bottomrule
\end{tabular}
    \caption{Word-based Evaluation Metrics under Free-form Answer Setting}
    \label{tab:f1}
\end{table}

\begin{table}[t]
    \centering
\begin{tabular}{lcccc}
\toprule
Metric & DMN & CMN & MQA & TM \\
\midrule
        \midrule
BERTScore (F1) & 87.20 & 88.15 & 88.16 & 87.33 \\
Sentence-BERT & 0.443 & 0.557 & 0.508 & 0.457 \\
        \bottomrule
    \end{tabular}
    \caption{Sentence-based Semantic Similarity under Free-form Answer Setting}

    \label{tab:bert}
\end{table}


\clearpage
\clearpage
\clearpage
\clearpage
\clearpage
\clearpage
\clearpage

\begin{table*}[t]
    \centering
    \begin{tabular}{p{0.25\linewidth} p{0.7\linewidth}}
        \toprule
        \textbf{Class} & \textbf{Tools / Techniques} \\
        \midrule
        System & Agroforestry, Chinampa, Crop rotation, Duck rice method, Floating Agriculture, Milpa System, Rice-fish system, Shade-grown coffee, Three Sisters method \\
        Landscape & Andes' steep, Below Sea Level Farming, Moray Circle Terrace, Polder, Terrace Farming, Xinghua Duotian \\
        Irrigation & Bisse d'Ayent, Clay Pot Irrigation, Drainage system, Furrow irrigation, Irrigation canal, Level Basin Irrigation, Noria, Pattern tile drainage, Qanat, Rainwater Harvesting Pit, Reservoir irrigation, Subak, Turpan Karez water system, Waterladder pump \\
        Soil & Composting, Controlled burning, Erosion barriers, Hugelkultur, Jhum Cultivation, Keyline design, Ridge and Furrow, Slash-and-burn, Sloping Agricultural Land Technology, Zai pits \\
        Processing & Basket press, Cotton gin, Flail, Grape-treading, Matcha Stone Mill, Metate, Mortar and pestle, Rice Sieving, Sheaf, Stook, Threshing machine, Treadmill, Watermill, Winnowing \\
        Tool & Digging stick, Foot plough, Hoe, Horse-ploughing, Machete, Ox-ploughing, Pitchfork, Rice Field Marker, Sickle, Wheelbarrow, Yoke \\
        Storage & Hasa Drying, High-floored storage, Karausu, Mangoku, Para-para drying rack, Tomi, Traditional fruit drying \\
        Pest & Botanical Pest Control, Scarecrows, Shishi-odoshi, Stork-friendly farming \\
        Practice & Hand Planting, Tea-picking \\
        \bottomrule
    \end{tabular}
    \caption{Categorized List of Traditional Methods (TM)}
    \label{tab:traditional_methods}
\end{table*}

\begin{table*}[h]
    \centering
    \begin{tabular}{p{0.25\linewidth} p{0.7\linewidth}}
        \toprule
        \textbf{Category} & \textbf{Machinery} \\
        \midrule
        Harvester & Aquatic Weed Harvester, Bean Harvester, Beet Harvester, Carrot Harvester, Coffee Bean Harvester, Combine, Corn Harvester, Daikon Radish Harvester, Forage Harvester, Grape Harvester, Harvester, Haulm Topper, Mechanical Tree Shaker, Onion Harvester, Peanut Harvester, Potato Harvester, Pumpkin Picking Machine, Reaper-Binder, Stripper, Sugarcane Harvester, Swather, Tomato Harvester, Yam Harvester \\
        Seeder & Autonomous Seeder, Broadcast Seeder, Seeding Machine, Semi-Automatic Seeder \\
        Transplanter & Cabbage Transplanter, Transplanter \\
        Planter & Planter, Potato Planter \\
        Handling & Bale Gripper, Bale Sledge, Claw Machine, Packing Machine, Sorting Machine \\
        Transport & Chaser Bin, Conveyor, Dump Trailer, Electric Cart, Gravity Wagon, Mother Bin \\
        Tillage & Cultivator, Disc Harrow, Drag Harrow, Excavator, Motorized Plough, Reversible Plough, Spading Machine, Spring-Tooth Harrow, Strip-Till Implement, Tiller \\
        Soil Preparation & Cambridge Roller, Destoner, Land Imprinter, Potato Bed Former, Puddling Machine, Ridge Maker, Ridging Hiller, Rotary Hiller, Smooth Roller \\
        Spreader & De-Icing Agent Spreader, Fertilizer Spraying Robot, Fertilizer Spreader, Manure Spreader, Silage Spreader, Sprayer \\
        Irrigation & Irrigation Machine, Sprinkler Irrigation, Water Wheel \\
        Chopper & Green Manure Chopper, Leaf Chopper, Root Cutting Machine, Shredder, Straw Shredder \\
        Baling & Roll Baler, Tree Baler \\
        Weeder & Tine Weeder, Weeder Machine \\
        Processing & Beet Cleaner Loader, Fanner, Rice Huller, Rice Milling Machine, Sugarcane Press, Threshing Machine \\
        Pruner & Grape Pruning Machine \\
        Rock Removal & Rock Windrower, Stone Picker \\
        Power Equipment & PTO \\
        Autonomous System & Smart Robotic Farmer, Unmanned Helicopter \\
        Facility Equipment & Multi-Tunnel, Vertical Farming System \\
        Planting Support & Border Coating Machine, Seed-Counting Machine, Seedling Machine \\
        Mower & Flail Mower, Mower \\
        \bottomrule
    \end{tabular}
    \caption{Categorized List of Machine QA (MQA)}
    \label{tab:machinery_categories}
\end{table*}

\clearpage
\clearpage
\clearpage
\clearpage
\clearpage
\clearpage
\clearpage
\clearpage

\end{document}